\documentclass[manuscript]{acmart}
\renewcommand\footnotetextcopyrightpermission[1]{} 
\copyrightyear{2025}
\acmYear{2025}
\acmConference[MM '25]{Proceedings of the 33rd ACM International Conference on Multimedia}{October 27--31, 2025}{Dublin, Ireland}
\acmBooktitle{Proceedings of the 33rd ACM International Conference on Multimedia (MM '25), October 27--31, 2025, Dublin, Ireland}
\acmDOI{10.1145/XXXXXXX.XXXXXXX} 
\acmISBN{979-8-4007-2035-2/2025/10} 

\usepackage{booktabs}
\usepackage{multirow}
\usepackage{graphicx}
\usepackage{hyperref}
\usepackage{array}

\usepackage{longtable}
\usepackage{multirow}
\usepackage{tabularx}
\usepackage{xcolor}

\usepackage{tikz}
\usetikzlibrary{positioning,arrows.meta,shapes.misc,fit,calc}
\usepackage{microtype}

\usepackage{subcaption}
\usepackage{chngcntr}
\usepackage{appendix} 
\usepackage{amsmath}

\title{RFM-HRI : A Multimodal Dataset of Medical Robot Failure, User Reaction and Recovery Preferences for Item Retrieval Tasks}

\author{Yashika Batra}
\affiliation{%
  \institution{Cornell University, Cornell Tech}
  \city{New York City, NY}
  \country{United States of America}}
\email{yb344@cornell.edu}

\author{Giuliano Pioldi}
\affiliation{%
  \institution{Cornell University, Cornell Tech}
  \city{New York City, NY}
  \country{United States of America}}
\email{gp433@cornell.edu}

\author{Promise Ekpo}
\affiliation{%
  \institution{Cornell University, Cornell Tech}
  \city{New York City, NY}
  \country{United States of America}}
\email{poe6@cornell.edu}

\author{Arman Sayatqyzy}
\affiliation{%
  \institution{Cornell University, Cornell Tech}
  \city{New York City, NY}
  \country{United States of America}}
\email{arman.sayatqyzy@gmail.com}

\author{Purnjay Maruur}
\affiliation{%
  \institution{Cornell University, Cornell Tech}
  \city{New York City, NY}
  \country{United States of America}}
\email{pm674@cornell.edu}

\author{Shalom Otieno}
\affiliation{%
  \institution{Cornell University, Cornell Tech}
  \city{New York City, NY}
  \country{United States of America}}
\email{soo26@cornell.edu}

\author{Kevin Ching}
\affiliation{
    \institution{Weill Cornell Medicine}
    \city{New York City, NY}
    \country{United States of America}}
\email{kec9012@med.cornell.edu}

\author{Angelique Taylor}
\affiliation{%
  \institution{Cornell University, Cornell Tech}
  \city{New York City, NY}
  \country{United States of America}}
\email{amt298@cornell.edu}

\authorsaddresses{}


\begin{abstract}
While robots deployed in real-world environments inevitably experience interaction failures, understanding how users respond through verbal and non-verbal behaviors remains under-explored in human–robot interaction (HRI). This gap is particularly significant in healthcare-inspired settings, where interaction failures can directly affect task performance and user trust. We present the Robot Failures in Medical HRI (RFM-HRI) Dataset, a multimodal dataset capturing dyadic interactions between humans and robots embodied in crash carts, where communication failures are systematically induced during item retrieval tasks.

\noindent Through Wizard-of-Oz studies with 41 participants across laboratory and hospital settings, we recorded responses to four failure types (speech, timing, comprehension, and search) derived from three years of crash-cart robot interaction data. The dataset contains 214 interaction samples including facial action units, head pose, speech transcripts, and post-interaction self-reports. Our analysis shows that failures significantly degrade affective valence and reduce perceived control compared to successful interactions.

\noindent Failures are strongly associated with confusion, annoyance, and frustration, while successful interactions are characterized by surprise, relief, and confidence in task completion. Emotional responses also evolve across repeated failures, with confusion decreasing and frustration increasing over time.

\noindent This work contributes (1) a publicly available multimodal dataset (RFM-HRI), (2) analysis of user responses to different failure types and preferred recovery strategies, and (3) a crash-cart retrieval scenario enabling systematic comparison of recovery strategies with implications for safety-critical failure recovery. Our findings provide a foundation for failure detection and recovery methods in embodied HRI.

\end{abstract}

\begin{document}


\maketitle
\section*{Acknowledgments}
\noindent This material was supported by the National Science Foundation under Grant No. IIS-2423127.






\section{Introduction}

The integration of autonomous systems into real-world deployments has highlighted the inevitability of interaction failure. 
When robots fail, humans inherently respond both verbally (i.e., speech \cite{giuliani_frontiers_nodate}) and non-verbally (i.e., gestures, facial expressions \cite{Das2021explainable}, posture etc).
However, the manner in which people respond to the robot may depend on the robot's use of verbal or non-verbal communicative cues during failures, how the robot chooses to recover from such failures, and how appropriate this response is. \cite{admoni_nonverbal_2016, ngo_human_2024}.
Despite this, limited work in HRI has explored methods by which robots can effectively recover from failures using multimodal cues in real-time \cite{giuliani_frontiers_nodate, stiber_not_2020}. 

To date, HRI research has primarily examined robot failures through asynchronous video evaluations \cite{parreira_study_2024} and in social environments \cite{mirnig_frontiers_nodate}, with limited attention to workplace settings where stressful conditions place different demands on users.
As a result, the HRI community could benefit from failure datasets that capture technical robot failures and user reactions related to safety critical scenarios.

Healthcare represents a particularly demanding safety-critical domain, characterized by extreme time pressure, rapidly changing conditions, and high cognitive load \cite{jacquet2018emergency}. During emergency procedures, delays or errors in accessing medications and supplies can disrupt workflow and divert attention away from patient care and team coordination, with outsized consequences for team performance and timely intervention \cite{calvo2025emergency}.

In emergency medicine, access to supplies is often mediated through shared physical artifacts such as crash carts, which centralize critical medications and equipment \cite{jacquet2018emergency}. Locating items on a crash cart is a distributed, collaborative activity that requires interpreting verbal requests, maintaining awareness of cart organization, and coordinating actions under stress \cite{taylor2025crashcart}. Furthermore, the frequency of crash cart use varies by hospital unit; as a result, some healthcare workers interact with the cart far less frequently and may struggle to recall where specific supplies are located. Our long-term vision is to design a crash cart robot that supports efficient access to critical supplies, reducing delays in patient care.

Within time-critical and coordination-intensive tasks, robot errors can be categorized from two perspectives: the technical system perspective, which focuses on deviations from the robot's designed functionality, and the user perspective, which centers on deviations from human expectations and impacts on interaction.
From the user perspective, failures tackle the problem of detecting robot errors using affect, gaze, or posture \cite{Frijns2024, honig2018understanding}. Complementary research models error detection from the system-perspective through norm violation or anomaly detection frameworks \cite{mottwe}.

While a growing body of work explores robot recovery behaviors, these efforts focus almost entirely on verbal repair and conversational agents \cite{alghamdi_system_2024}.
Datasets such as the ERR@HRI 2.0 Challenge \cite{spitale2024err,cao2025err}, REFLEX \cite{khanna2025reflex}, REACT \cite{candon2024react}, Response-to-Errors \cite{stiber2023using}, and EMPATHIC \cite{cui2021empathic} have propelled research interest in this domain, by providing multimodal recordings of conversations, social signals, and user reactions.

Despite this progress in socially grounded and conversational settings, several gaps remain.
First, because many robot failure modes studied in HRI are designed for social settings, their applicability to safety-critical workplace environments, such as healthcare, remains unclear.
Second, existing HRI datasets do not provide paired annotations of failure events and user-preferred recovery behaviors in physically grounded healthcare interactions.
Third, even when recovery strategies are studied, their preference structure across distinct failure types has not been systematically examined in safety-critical contexts.

\begin{figure}[t]
\centering
\includegraphics[width=1.0\columnwidth]{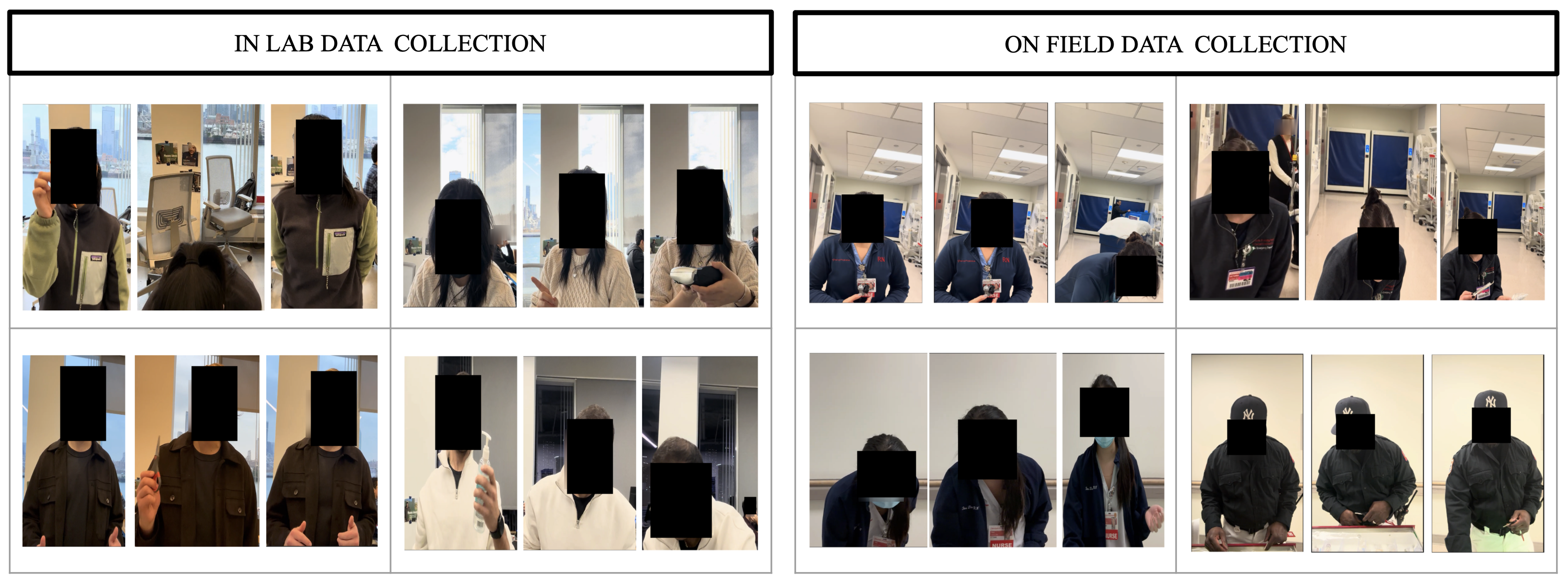}
\caption{Data collection environment at university campus laboratory (left) and at the hospital (right) Participants faces are blacked out in compliance with our approved IRB protocol. \color{black}}
\label{fig:data_collection_env}
\end{figure}

These challenges motivate the study of assistive robotic systems for rapid, reliable item retrieval in healthcare-oriented tasks. In this paper, we address these gaps by conducting a structured failure-induction study with a crash cart robot that guides users through item retrieval tasks.
As part of this study, we first derived four failure types (speech, timing, comprehension, and search) from three years of empirical observation of interactions with crash cart robots in emergency medicine contexts.
We then collected a multimodal dataset including 214 samples of HRI in laboratory and hospital environments, which capture synchronized facial expressions, head pose, speech, and post-interaction reflections on recovery strategy preferences (Figure \ref{fig:data_collection_env}).

The item retrieval task captures core elements of general item retrieval problems in embodied robotics, including goal specification, spatial search, manipulation, communication, and recovery from failure. These challenges are shared across a range of real-world deployments, such as manufacturing and factory settings, where robots must reliably retrieve, transport, and hand off objects under time pressure and partial observability. As such, RFM-HRI can support research on item retrieval, failure detection, and recovery strategy learning beyond healthcare domains.

Our \textbf{contributions are threefold}.
\textbf{First}, we introduce RFM-HRI, a novel multimodal dataset of dyadic interactions with a crash cart robot, which will be made publicly available to increase research interest in evaluating robot failures and recovery strategies. The dataset has 214 samples from 41 participants across laboratory and hospital settings, each including synchronized facial action units, head pose, speech transcriptions, and post-interaction self reports of emotional reactions and robot recovery preferences.
\textbf{Second}, we design and evaluate a set of robot communication failure modes, including speech, timing, search, and comprehension failures, and analyze how users respond to each failure type and its associated robot recovery strategy.
Failure trials are associated with more negative emotional responses and reduced sense of control relative to successful trials, with confusion, annoyance, and frustration dominating reported emotions. Emotional patterns vary across failure types and over repeated interactions. We also document a strong preference for verbal recovery strategies, with multimodal and nonverbal strategies selected far less frequently. 
\textbf{Third}, we contribute a novel HRI failure scenario centered on guided search and item retrieval tasks in a crash cart robot, providing a realistic interaction for studying failure and recovery behavior in healthcare and other structured item retrieval settings.
Through a systematic comparison of verbal and non-verbal robot recovery strategies, we offer design implications for robots that engage in failure to recovery interaction loops in safety-critical HRI settings.

\section{Related Work}

\subsection{Conceptual Foundations}
An error is the internal manifestation or state change resulting from the fault (e.g., the robot's navigation system calculates an incorrect internal position).
A failure is the ultimate external event where the system deviates from its expected service, becoming observable to the human user (e.g., the robot bumps into a wall or violates a social norm by interrupting) \cite{laprie1995dependable}. 

\subsubsection{Definition of Robot Failures}
\label{sec:robot_errors}

Based on the observations of the robot errors from robot field studies over the course of 3 years \cite{taylor2025crashcart, tanjim2025help, tanjim_human-robot_2025},  we identified four communication errors that occur during HRIs in the emergency room. Before describing these specific failures, we clarify two complementary perspectives for understanding robot errors in HRI.

\subsubsection{Error Perspectives}

We distinguish between two perspectives for analyzing robot failures, system and user perspective errors.

\emph{System Perspective Errors} are failures or undesired behaviors that originate from the robot's internal components or algorithms, such as a sensor malfunction or delayed speech output. These are considered robot-side faults when viewed from the developer's perspective, and represent technical breakdowns in the robot's designed functionality.

\emph{User Perspective Errors} are failures that are experienced by the human user due to a mismatch, miscommunication, or human misunderstanding. These are failures that are perceived in the context of HRI, even if the robot hardware or software is technically operating correctly. User perspective errors capture how failures manifest in the lived experience of interaction, regardless of their technical root cause.

Differentiating between these perspectives explicitly allows us to label and analyze failure events in terms of both internal root causes (system perspective) and human experience (user perspective). This dual framing is essential for developing robots that can both detect technical malfunctions and recognize when users perceive something has gone wrong.
This terminology provides a shared vocabulary for connecting technical breakdowns to how failures are perceived and managed during interaction, which motivates prior HRI models of failure handling.

\subsection{Robot Failures}
In HRI, robot failures are increasingly treated as interaction events that shape what users notice, how they interpret robot behavior, and what corrective actions they attempt \cite{honig2018understanding}.
Honig and Oron-Gilad synthesized prior work on failure handling and introduced the Robot Failure Human Information Processing (RF-HIP) model, which frames failure handling as staged human information processing (e.g., failure communication, perception and comprehension, and problem solving), with each stage influenced by contextual factors and mitigation strategies \cite{honig2018understanding}.
They further argued that failure handling has downstream consequences for trust, performance, and willingness to reuse robots, motivating datasets that capture not only failures themselves but also how users experience and respond to them \cite{honig2018understanding}.

These concerns are further sharpened in healthcare and emergency settings, where time pressure, safety criticality, and team coordination demands can amplify the consequences of failures \cite{taylor2025crashcart}. Taylor et al.\ presented a rapidly built medical crash cart robot deployed in emergency rooms, and introduced a taxonomy of failure modes observed during real-world clinical use, including breakdowns in communication, timing, and coordination. This work highlights how robot failures in healthcare are embedded within broader team workflows and clinical constraints, and underscores the importance of analyzing failure and recovery in embodied medical robots \cite{taylor2025crashcart}. 

To date, healthcare robot deployments have largely emphasized qualitative analysis and system-level insights, and fewer efforts have focused on releasing publicly available datasets that combine labeled failure modes with multimodal annotations of user reactions for machine learning research \cite{taylor2025crashcart}.

Our work complements this body of research by pairing structured failure induction with multimodal measurements of user responses in a controlled experimental setting, supporting both human-centered analysis and data-driven modeling of human reactions to robot failures.

\subsection{Robot Error Detection Using Social Signals}
A complementary line of work studies how \emph{social signals}, including facial expressions, head movements, gaze, prosody, and body posture, can be used to recognize when a robot-assisted task is going wrong \cite{bremers2023socialcues}. Bremers et al.\ review this literature and argue that social cues can be early indicators of task failure, occurring before explicit verbal complaint or task abort \cite{bremers2023socialcues}. They highlight that models leveraging gaze shifts, facial tension, hesitation markers, or changes in speaking rate can distinguish successful from failed interactions, while also noting that many existing studies are small-scale and domain-specific \cite{bremers2023socialcues}.

Recent multimodal work operationalizes this idea in conversational settings \cite{cao2025err}. The ERR@HRI 2.0 challenge provides a dataset of dyadic human-robot conversations with LLM-powered robots, annotated for robot errors and users' corrective intent, with accompanying multi modal features\cite{cao2025err}. This corpus supports automatic error detection from facial and vocal cues, but its tasks (e.g., trip planning, debates) differ substantially from time-critical medical retrieval and physically grounded human-robot coordination \cite{cao2025err}.

Most similar to ``social-signal response to errors'' in physical HRI, the \emph{Response-to-Errors in HRI} dataset curated by Stiber et al.\ aggregates multiple physical HRI scenarios and annotates temporal boundaries around robot errors and human reactions, with facial action units extracted via OpenFace \cite{stiber2023using}. In their setup, each entry corresponds to an error episode (pre-error, during-error, reaction, post-error) with annotated timestamps for perceived error start and reaction onset/offset \cite{stiber2023using}. This setup provides an important precedent for treating reactions as temporally localized signals around induced or unexpected errors in embodied tasks \cite{stiber2023using}.

Our dataset follows the social-signal perspective but in a different setting. We study an embodied crash cart robot that uses sound and lights to guide item retrieval during simulated emergency scenarios, where failures are systematically injected across speech, timing, comprehension, and search \cite{bremers2023socialcues}. We record videos, including facial expressions
 (processed with Google MediaPipe-style landmark pipelines), audio (processed with speech-to-text), and post-study survey responses after participants re-watch their interactions, allowing us to relate objective social-signal patterns to both online behavior and retrospective self-reports in a physically grounded task \cite{bremers2023socialcues}.

To situate these signals and failure annotations within the broader landscape of available resources, we next summarize relevant datasets spanning affective computing, conversational error modeling, and embodied HRI.

\begin{table}[t]
\scriptsize
\centering
\setlength{\tabcolsep}{3pt}
\renewcommand{\arraystretch}{1.2}
\caption{Comparison of representative datasets spanning facial expression analysis, conversational errors, and human–robot interaction. In contrast to prior work, \textbf{RFM-HRI (Ours)} provides a healthcare-centered HRI dataset featuring a novel crash cart robot, controlled failure injection, and rich multimodal annotations of human reactions, including speech, FACS and AUs, head pose, gaze estimation, and post-hoc survey responses—collected from both expert and non-expert participants.}
\resizebox{\columnwidth}{!}{
\begin{tabular}{
p{1.3cm}
p{3.0cm}
p{2.0cm} 
>{\centering\arraybackslash}p{1.1cm} 
>{\centering\arraybackslash}p{1.4cm} 
>{\centering\arraybackslash}p{1.4cm} 
>{\centering\arraybackslash}p{1.1cm} 
>{\centering\arraybackslash}p{1.1cm} 
}
\toprule
\textbf{Dataset} &
\textbf{Annotations} &
\textbf{Dataset Modalities} &
\textbf{Healthcare Domain?} &
\textbf{Experts and Non-Experts?} &
\textbf{Human Robot Interaction?} &
\textbf{Controlled Failure Injection?} &
\textbf{Reactions to Failures?} \\
\midrule

Oops! \cite{epstein2020oops} &
Videos Annotated for Action Intention Variability &
Video &
$\times$ &
$\times$ &
$\times$ &
$\times$ &
$\checkmark$
 \\
 
RAF-DB \cite{li2017reliable} &
Compound Expression Images and Labels in the Wild &
Image &
$\times$ &
$\times$ &
$\times$ &
$\times$ &
$\times$
\\

CK+ \cite{lucey2010extended} &
Facial Action Coding System, Action Units, and Labels for Posed and Unposed Expressions &
Video, Image, FACS/AUs &
$\times$ &
$\times$ &
$\times$ &
$\times$ &
$\times$
\\

Oulu-CASIA \cite{zhao2011facial} &
Facial Expression Dataset, Across Illumination Conditions &
Image (Infrared and Visible Light) &
$\times$ &
$\times$ &
$\times$ &
$\times$ &
$\times$
\\

AFEW 7.0 \cite{dhall2012collecting} &
Emotion-Labeled Movie Clips &
Video, Audio &
$\times$ &
$\times$ &
$\times$ &
$\times$ &
$\times$
\\

BP4D-Spontaneous \cite{zhang2013high} &
2D and 3D Dataset Containing Facial Action Units, Video Features, and Head Poses &
Video (2D and 3D), FACS/AUs, Pose &
$\times$ &
$\times$ &
$\times$ &
$\times$ &
$\times$
\\

ACE \cite{aneja2020conversational} &
Conversational Errors and Analysis in Human-Agent Interaction &
Transcripts, Annotations &
$\times$ &
$\times$ &
$\times$ &
$\checkmark$ &
$\checkmark$
\\

ERR@HRI 2.0 \cite{cao2025err} &
Multimodal Dataset of Conversational HRI with System-Failures &
FACS/AUs, Pose, Audio Features, Transcripts &
$\times$ &
$\times$ &
$\checkmark$ &
$\checkmark$ &
$\checkmark$
\\

EMPATHIC \cite{cui2021empathic} &
Implicit human feedback (facial reactions) &
Video, Annotations &
$\times$ &
$\times$ &
$\times$ &
$\checkmark$ &
$\checkmark$
\\

REACT \cite{candon2024react} &
Human reactions \& evaluative feedback &
FACS/AUs, Pose, Gaze, Annotations &
$\times$ &
$\times$ &
$\checkmark$ &
$\checkmark$ &
$\checkmark$
\\

REFLEX \cite{khanna2025reflex} &
Human reactions to failures/explanations &
FACS/AUs, Pose, Gaze, Transcript, Annotations &
$\times$ &
$\times$ &
$\checkmark$ &
$\checkmark$ &
$\checkmark$
\\

Response-to-Errors \cite{stiber2023using} &
Facial AU social responses to errors &
AUs, Annotations &
$\times$ &
$\times$ &
$\checkmark$ &
$\checkmark$ &
$\checkmark$
\\

HRI-SENSE \cite{gucsi2025hri} &
Multimodal social/emotional responses &
Videos (with Depth), FACS/AUs, Pose, Gaze, Transcripts, Survey Data &
$\times$ &
$\times$ &
$\checkmark$ &
$\checkmark$ &
$\checkmark$
\\

OpenRoboCare \cite{liang2025openrobocare} &
Expert caregiving tasks &
Video (RGB--D), Pose, Gaze, Annotations, Tactile Data &
$\checkmark$ &
$\times$ &
$\checkmark$ &
$\times$ &
$\times$
\\

\hline

\textbf{RFM-HRI (Ours)} &
\textbf{Robot Failures (and Recovery) in Medical HRI} &
\textbf{FACS/AUs, Pose, Gaze, Transcripts, Survey Data} &
$\checkmark$ &
$\checkmark$ &
$\checkmark$ &
\textbf{$\checkmark$} &
\textbf{$\checkmark$} \\
\end{tabular}}

\label{tab:datasets}
\end{table}

\subsection{HRI Datasets for Failures and Failure Responses}
We summarize datasets that capture unintentional events and affect, conversational errors, and embodied HRI failures and reactions, including healthcare-focused multimodal resources (Table~\ref{tab:datasets}).

\subsubsection{Unintentional Events and Affective Computing.}
The OOPS! dataset collects in-the-wild videos of unintentional human actions and labels the transition from intentional behavior to accidental ``oops'' moments \cite{epstein2020oops}. It is a valuable resource for learning to anticipate accidents, but it does not involve robots, healthcare settings, or explicit user reactions to robot behavior \cite{epstein2020oops}.

Large facial-expression datasets such as RAF-DB \cite{li2017reliable}, CK+ \cite{lucey2010extended}, Oulu-CASIA \cite{zhao2011facial}, AFEW 7.0 \cite{dhall2012collecting}, BP4D-Spontaneous \cite{zhang2013high}, and related corpora provide labels for emotions and/or Facial Action Coding System (FACS) and Action Units (AUs), across posed and in-the-wild (or semi-structured) recordings. 
For example, Oulu-CASIA is explicitly designed to include both near-infrared (NIR) and visible recordings for basic facial expressions \cite{zhao2011facial}, while BP4D-Spontaneous provides high-resolution dynamic facial behavior with frame-level FACS annotations and 3D facial and head tracking metadata, captured under a structured elicitation protocol \cite{zhang2013high}. These datasets underpin many tools for AU detection and affect modeling, but they are not collected in HRI settings and do not target reactions to task or safety-critical robot failures.

\subsubsection{Conversational Error Datasets.}
Within conversational AI, the ACE dataset annotates conversational agent errors in human-agent interactions to support the analysis of error types and user-perceived breakdowns \cite{aneja2020conversational}.
ERR@HRI-style corpora extend beyond text to include multimodal behavioral features and temporally localized error/reaction annotations in conversational HRI \cite{cao2025err}.
However, these datasets focus on conversational tasks and do not capture physical interactions with robots during urgent medical item retrieval tasks \cite{aneja2020conversational, cao2025err}.

\subsubsection{Embodied HRI Failures and Reactions.}
Within HRI and human-agent interaction, multiple datasets explicitly connect agent/robot mistakes to human reactions.
The EMPATHIC \cite{cui2021empathic} framework includes a dataset of human facial reactions while participants \emph{observe} an autonomous agent that executes (often suboptimal) behavior, with structured incentives so that observers care about outcomes. This is valuable for learning reaction-to-quality mappings, but it is primarily observational rather than a physically interactive, time-critical collaboration setting \cite{cui2021empathic}.
The REACT \cite{candon2024react} database contributes two data sets that capture user natural nonverbal reactions during human-robot interactions over time (e.g., a collaborative game and a photography scenario) and includes explicit evaluative feedback signals alongside nonverbal features and contextual traces \cite{candon2024react}. 
REFLEX \cite{{khanna2025reflex}} extends the ``reaction to failure'' framing by capturing human reactions not only to induced robotic action failures but also to different explanation levels and strategies across repeated interactions, with dataset organization and modalities described in dedicated dataset sections \cite{khanna2025reflex}. 
Finally, HRI-SENSE provides a multimodal time-synchronized data set of social/physical behaviors and self-assessed questionnaires (e.g., frustration, satisfaction) during collaborative manipulation with a TIAGo robot, emphasizing multimodal signals and subjective impressions in embodied interactions \cite{gucsi2025hri}.

\subsubsection{Healthcare-Focused Multimodal Datasets.}
Finally, multimodal datasets focused on healthcare care, such as OpenRoboCare, demonstrate the value of rich sensing for care.
OpenRoboCare provides expert occupational therapist demonstrations of Activities of Daily Living and spans five modalities (RGB-D video, pose tracking, eye-gaze tracking, tactile sensing, and task/action annotations) to study physical caregiving routines \cite{liang2025openrobocare}.
The dataset is designed around expert demonstrations and caregiving principles rather than systematically induced robot failures and observed user reactions under time pressure \cite{liang2025openrobocare}.

Taken together, these resources motivate the need for datasets that jointly capture structured failures and multimodal human responses in high-stakes embodied tasks. Relative to prior work, our dataset is distinguished by (i) a healthcare-motivated \emph{crash cart robot} guiding item retrieval, (ii) \emph{structured, labeled failure injection} spanning speech/timing/comprehension/search, and (iii) aligned multimodal social signals \emph{plus} post-hoc reflection after participants re-watch their own interaction \cite{taylor2025crashcart}.

\section{User Study: Medical Human-Robot Interaction Failures}

\subsection{Task Overview, Robotic Platform, and Participants}
\subsubsection{Task Overview: Crash Cart Item Retrieval}
In emergency and acute-care settings, clinicians frequently rely on crash carts to access medications and equipment during time-critical procedures \cite{taylor2025crashcart}. Locating supplies on a cart can involve distributed coordination and decision-making under time pressure, as clinicians navigate equipment organization and communicate amid uncertainty \cite{calvo2025emergency}. Breakdowns in this process, such as delays in locating required items or miscommunication, have been linked to poorer procedural coordination and are especially consequential when rapid response is needed, motivating careful organization and rapid access to supplies \cite{jacquet2018emergency}. To study how assistive robotic systems might support this form of situated, high-stakes collaboration, we focus on the task of guided item retrieval from a crash cart, using it as a representative interaction scenario in which timing, clarity of guidance, and shared understanding are essential.

\subsubsection{Robotic Platform}
\begin{figure}[t]
\centering
\includegraphics[width=1.0\columnwidth]{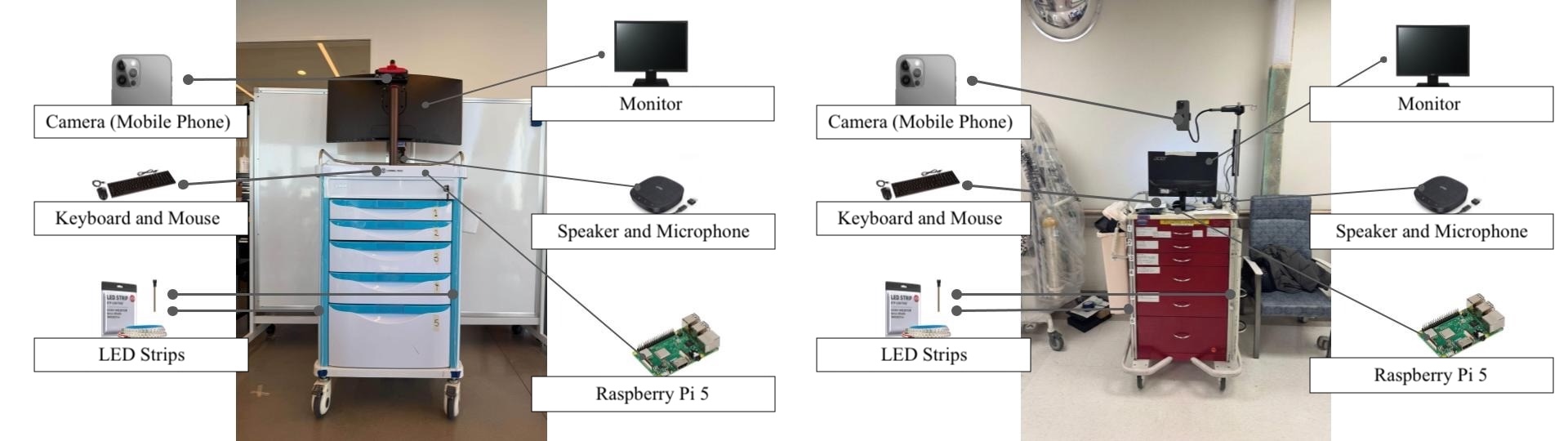}
\begin{minipage}{0.48\textwidth}
\centering
\small
\textbf{In Lab Data Collection}
\end{minipage}
\hfill
\begin{minipage}{0.48\textwidth}
\centering
\small
\textbf{In Field Data Collection}
\end{minipage}
\caption{Robotic crash cart platforms in real-world hospital deployment (left) and controlled university campus laboratory evaluation (right). \color{black}}
\label{fig:robot_platform}
\end{figure}

In our work, we employ a semi-autonomous crash cart robot designed to support item retrieval tasks in healthcare settings, serving as the primary apparatus for the \textit{RFM-HRI} dataset collection. Figure~\ref{fig:robot_platform} showcases the assembled carts, both in our university and hospital setup.
The platforms are built by retrofitting a standard crash cart to operate as a robot with multimodal cues, aligned with prior work in high-stakes healthcare settings, enabling realistic placement and interaction in clinical environments. 
To preserve validity during real-world hospital deployment, the drawers and shelves were stocked with standard medical supplies; however, for the controlled laboratory evaluation, we utilized distinct proxy items (e.g., office and hardware supplies) to simulate retrieval tasks. 
Table~\ref{tab:drawer_organization} details the specific inventory comparisons between the hospital and laboratory configurations.
Visual cues are provided by addressable LED strips mounted alongside the drawer columns. Each column is associated with and identified by an LED segment, allowing the robot to highlight spatial locations using light cues. The robot also includes a verbal interface that uses a text-to-speech (TTS) engine to deliver standard prompts, status updates, and guidance messages. User speech is captured through an onboard microphone and later transcribed manually or via automatic speech recognition (ASR) with correction for downstream analysis. 

The robot additionally incorporated an egocentric camera to capture first-person video of the interaction. Due to the distributed nature of the study, we did not enforce standardized camera hardware or video resolutions; however, the visual features reported in the dataset are standardized during post-processing (see Section~\ref{sec:facial_head_pose_features}).

To support post-hoc annotation and analysis, the robot maintained internal system logs indicating the start and end of each trial, as well as the specific scripted communication failure type (speech, timing, search, or comprehension) induced on each trial. While we only induced technical (system-level) failures, during interaction, the robot can experience failures from both technical (system-level) and experiential (user-level) perspectives. We define these failure types in Section \ref{sec:robot_errors}.

\begin{table*}[t]
\caption{Drawer-by-drawer inventory comparison of crash cart contents in the lab prototype vs hospital reference cart, supporting multimodal data collection and failure-induction studies.}
\resizebox{\textwidth}{!}{
\begin{tabular}{l|l|l}
\hline
\multicolumn{1}{c|}{\textbf{Drawer \#}} &
\multicolumn{1}{c|}{\textbf{Hospital Data Collection Supplies}} &
\multicolumn{1}{c}{\textbf{In-Lab Data Collection Supplies}}                                                                                                                                                            \\ \hline
Drawer 1                                
& \begin{tabular}[c]{@{}l@{}}Phenylephrine HCI, Epinephrine Injection, Metoprolol, \\ Naloxone, Nitroglycerin, Norepinephrine, Adenosine\end{tabular}
& Blue tape, white tape, SD cards, LEDs \\

Drawer 2
& \begin{tabular}[c]{@{}l@{}}Sodium Bicarbonate Injection USP, Lidocaine, \\ 50\% Dextrose Injection, 10\% Calcium Chloride Injection\end{tabular}
& \begin{tabular}[c]{@{}l@{}}Orange pencils, yellow pencils, batteries, \\ measuring tape\end{tabular} \\

Drawer 3
& \begin{tabular}[c]{@{}l@{}}Razor, Medical Tape, Surgical Gloves, Line Draw Arterial \\ Blood Sample Syringe\end{tabular}
& Red pencils, wires, blue markers, blue pencils \\

Drawer 4
& Blanket, Scissors, Face Masks
& \begin{tabular}[c]{@{}l@{}}White board cleaner, circuit board, calculator, \\ gloves, playstation controller, mouse\end{tabular} \\

Drawer 5 
& 1 mL Tuberculin Syringe, 6 mL Syringe, 500 mL IV Bag 
& \begin{tabular}[c]{@{}l@{}}Scissors, pliers, book, wipes, orange multimeter, \\ hand sanitizer\end{tabular} \\

Drawer 6
& Drawer 6 is Empty in the Hospital Cart
& \begin{tabular}[c]{@{}l@{}} No Drawer 6 on In-Lab Cart \end{tabular} \\ \hline
\end{tabular}
}
\label{tab:drawer_organization}
\end{table*}

\subsubsection{Participants}
We recruited 41 participants for an IRB-approved study through university campus advertisements and targeted outreach to Emergency Medicine listservs of Registered Nurses and medical residents in the Global North using convenience and purposive sampling. Of the 41 participants, 29 participated in lab studies and 12 in hospital studies. All participants completed a 30-minute study session and were compensated with a \$10 gift card.
Demographic information was collected voluntarily. 39\% provided complete responses. Among those who responded, the sample included 11 male and 5 female participants, with ages ranging from 22 to 44 years (M = 25.8, SD = 6.4).
Participants consisted of two groups, including 12 healthcare professionals and 29 non-experts. Participants in healthcare included Registered Nurses and Residents. 
Among healthcare workers, years of clinical experience ranged from 10 to 15 years. Lay participants were recruited from the general university population and reported their educational attainment, including Master's and doctoral/professional degrees.
Participants were not required to have prior experience with robotic systems.


\subsection{Experimental Methodology}
\subsubsection{Study Design} \label{sec:study_design}
We conducted a within-subjects study with five conditions to examine how participants respond to robot failures during a crash cart item retrieval task. Each participant experienced all five conditions in randomized order, including four failure conditions (speech failure, timing failure, comprehension failure, and search failure) and one success condition (error-free interaction). Study sessions were scheduled for approximately 20 minutes, with first 7-8 minutes dedicated to interaction trials with the crash cart robot. Due to time constraints, a small number of participants completed slightly fewer or more than five trials; this introduces minor imbalance across conditions, which is accounted for in our analyses.

Participants completed a predetermined sequence of five trials, four of which included injected robot failures and one of which was error-free, presented in randomized order. This design allowed us to compare participant behavior and subjective experience across different types of failures while controlling for individual differences.

The study centered on a guided search and retrieval scenario in which the robot provided spoken instructions and visual clues to direct participants to specific items in the crash cart drawers. Failures were introduced systematically during interaction trials, (see Section \ref{sec:failure_injection}). Participants were instructed to follow the robot’s instructions as naturally as possible and were not informed in advance that failures would occur.

\subsubsection{Study Setting}

We conducted data collection in both hospital and laboratory settings to balance ecological validity with practical constraints on participant recruitment. In-situ studies with clinical professionals, particularly in emergency care contexts, are inherently limited by time pressure and extended shifts, making sustained access to emergency room nurses infeasible. To complement hospital-based data and enable broader participation, we therefore replicated the core retrieval task in a controlled laboratory environment.
The laboratory setting featured a crash cart robot stocked with proxy items (Table \ref{tab:drawer_organization}), allowing systematic study of item retrieval behaviors with non-clinician participants while preserving task structure and interaction dynamics. In the hospital setting, the robot was deployed in a clinical training area and stocked with standard medical supplies (Table \ref{tab:drawer_organization}) to maintain realism.
Together, this dual-setting design supports analysis across realistic clinical contexts and scalable laboratory conditions while accounting for real-world constraints on clinical participation.

\subsubsection{Wizard-of-Oz Interface and Control}
\label{sec:woz_interface}
\begin{figure}[t!]
    \centering
    \begin{subfigure}[b]{\columnwidth}
        \centering
        \includegraphics[width=\columnwidth]{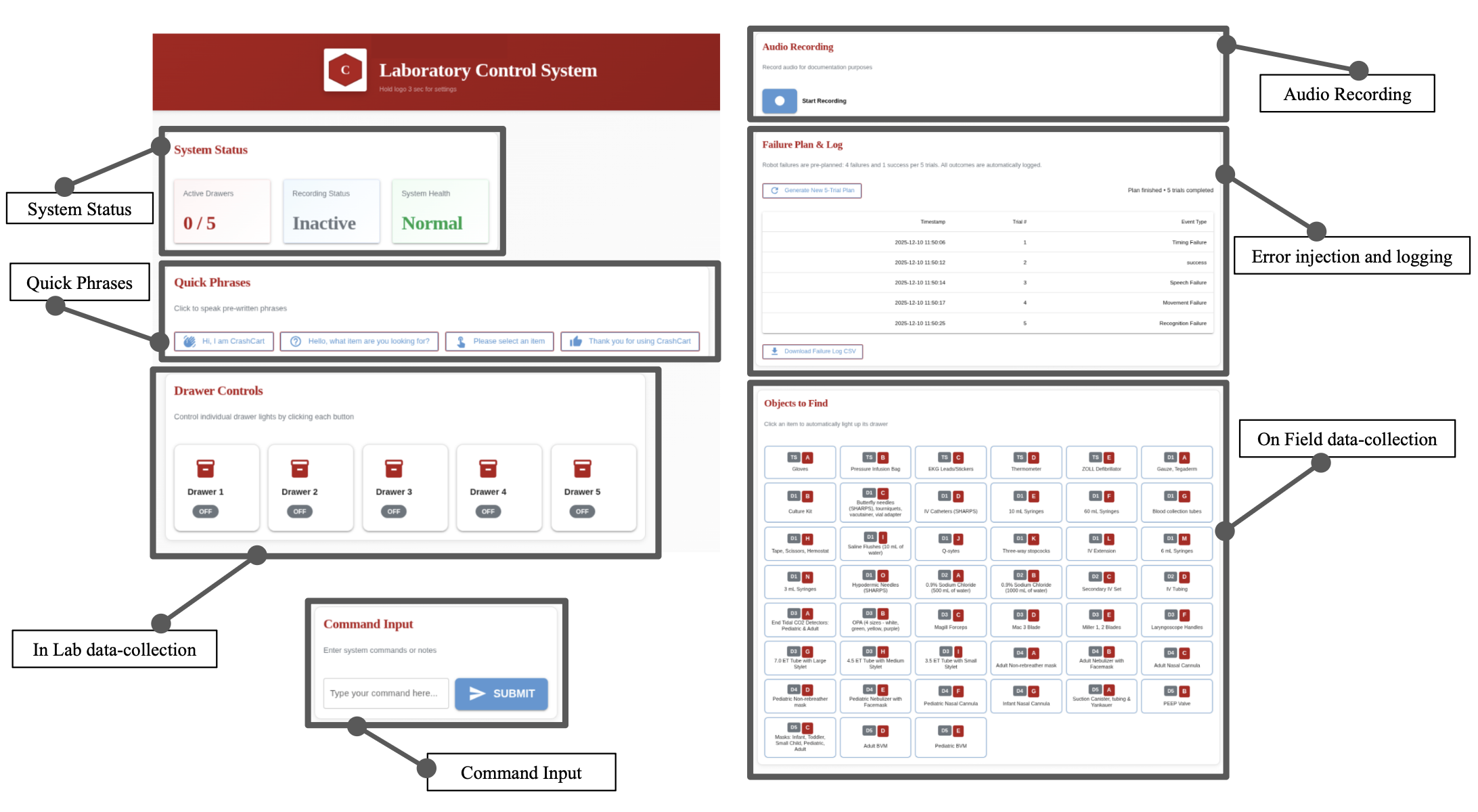}
        \caption{\textbf{Interface Overview:} The modular GUI features areas for monitoring, speech, and inventory.}
        \label{fig:woz_overview}
    \end{subfigure}
    \caption{The Wizard-of-Oz (WoZ) Interface components (Part 1 of 4).}
    \label{fig:woz_part1}
\end{figure}

To facilitate real-time control during the study, we developed a custom web-based Graphical User Interface (GUI), shown in Figure~\ref{fig:woz_overview}, which served as the primary control surface for the Wizard-of-Oz operator. 
This interface enabled the trained WoZ operator to manage multimodal robot behaviors, including addressable LED activation (Figures~\ref{fig:woz_lab_manual} and \ref{fig:woz_hospital_grid}), text-to-speech generation via pre-set ``Quick Phrases'' (Appendix Figure~\ref{fig:woz_status}) custom text input (Appendix Figure~\ref{fig:woz_audio}), and audio data logging (Appendix Figure~\ref{fig:woz_audio}).

\begin{figure}[t!]
    \begin{subfigure}[b]{\columnwidth}
        \centering
        \includegraphics[width=0.9\columnwidth]{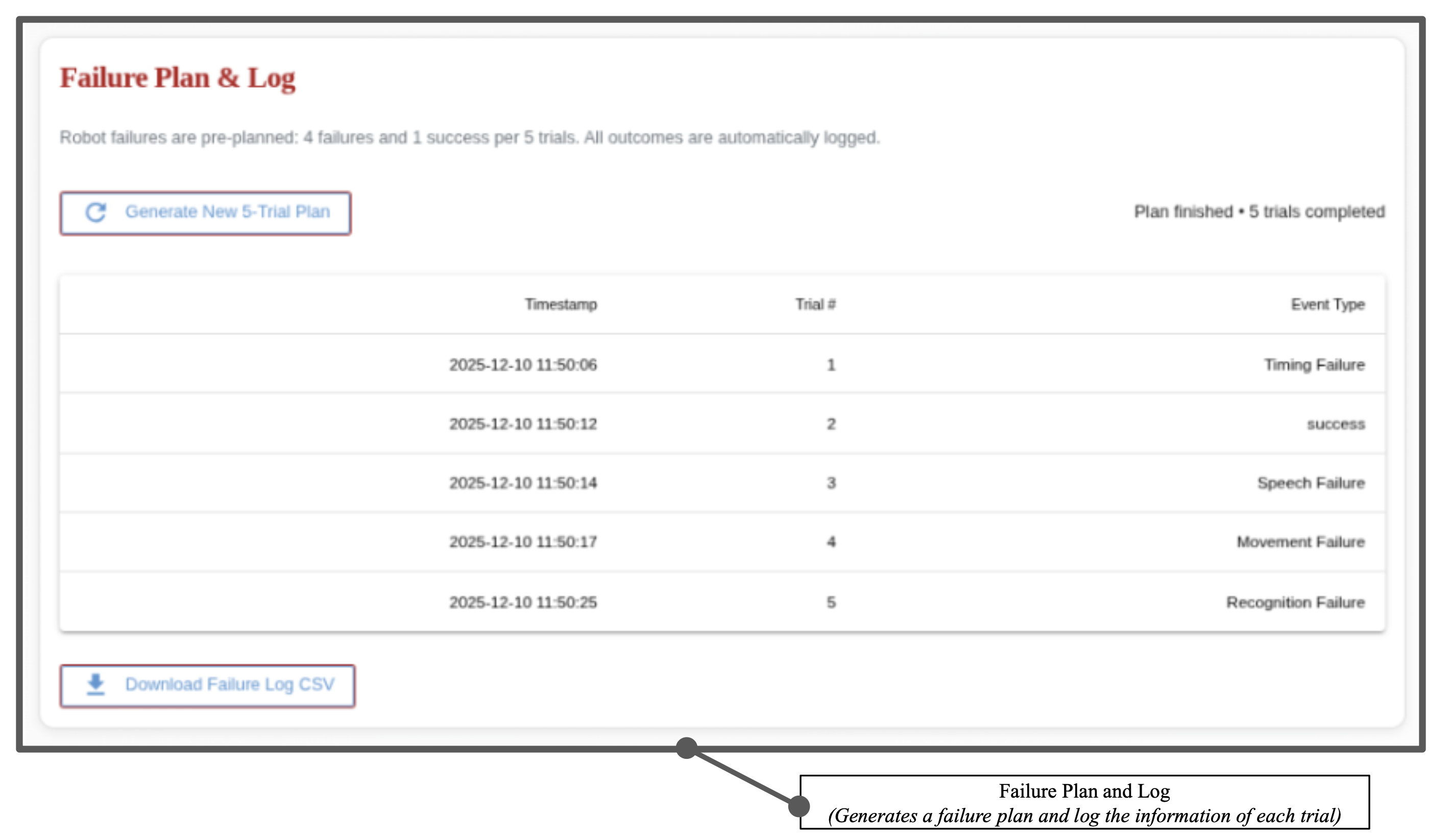}
        \caption{\textbf{Failure Injection Log:} Sequences failures and logs trial outcomes.}
        \label{fig:woz_failure}
    \end{subfigure}

    \caption{The Wizard-of-Oz (WoZ) Interface components (Part 2 of 4).}
    \label{fig:woz_part2}
\end{figure}

The interface organization was adapted to accommodate the differing inventory complexity between settings. 
In the laboratory evaluation, where the item set was limited, the operator utilized an external physical reference table to map requested items to their corresponding spatial locations, manually activating the specific drawer controls shown in Figure~\ref{fig:woz_lab_manual} (e.g., toggling "Drawer 1"). 
Conversely, for the on-field hospital data collection, the high volume of medical supplies required a more streamlined approach. We therefore implemented a dedicated "Objects to Find" module within the GUI (Appendix Figure~\ref{fig:woz_hospital_grid}), mapping specific medical items (e.g., "EKG Leads", "Gloves") to their hardware addresses. This allowed the operator to select the requested object directly from a digital grid, automatically triggering the correct drawer and LED cues without the latency of manual lookup.
Finally, the interface included the failure injection controls (Figure~\ref{fig:woz_failure}), allowing the operator to trigger the system anomalies defined in Section~\ref{sec:robot_errors} according to the study protocol.

Using this interface, we conducted a controlled Wizard-of-Oz (WoZ) study to investigate HRI dynamics during a clinically inspired object–retrieval task. Throughout all sessions, participants were led to believe that the robot operated autonomously. The WoZ operator was stationed in close proximity to the robot, positioned directly behind or adjacent to the cart, and controlled the system using the platform's onboard monitor, mouse, and keyboard. 
To maintain the illusion of autonomy despite this proximity, the operator's presence was framed as supervisory (e.g., monitoring system logs) rather than active control.
Although the system currently relies on human-in-the-loop control, the WoZ protocol was designed to mirror the behavioral logic of a future autonomous crash cart assistant. The protocol follows best practices in Wizard-of-Oz HRI research and aligns with recent healthcare-focused robotic field deployments \cite{tanjim2025help,taylor2025crashcart}.

Concretely, the WoZ operator executed all robot behaviors during the retrieval task using a standardized interaction script. Each interaction began with a standardized greeting by the WoZ: “Please retrieve X item”. The participants responded verbally by asking the cart for the desired item. The WoZ operator then selected the corresponding item from a predefined inventory list, which was mapped internally to the physical storage location of the item. Upon selection, the cart generated a multimodal response: it verbally announced the target drawer or shelf and simultaneously illuminated the LED segment corresponding to the correct drawer. Items stored on the open top shelf were indicated through speech alone. Once the participant retrieved the item, the operator deactivated the visual cues and optionally activated a closing message (e.g., “Thanks for using the Crash Cart”) to clearly signal task completion.
At no point were the failure conditions revealed to participants; if a participant questioned whether the robot was malfunctioning, the facilitator responded using a neutral, scripted reply to avoid biasing interpretation.

\subsubsection{Failure Injection} \label{sec:failure_injection}
We developed a dedicated failure injection module within the Wizard-of-Oz interface (previously described in Section \ref{sec:woz_interface}) that triggers controlled breakdowns in robot behavior in terms of system perspective errors (see Section \ref{sec:robot_errors}) during the drawer retrieval task. The module introduces four failure types that reflect common sources of miscoordination in clinical HRI. Each failure is executed through a function during live interaction.

The system failures are further explained below:

\begin{itemize}

\item \emph{Speech Failure.} The robot provides an under-specified or incomplete instruction, by telling the participant to “Open the drawer.” without specifying which drawer to open. For context, there are five drawers on the carts used for this study. Hence, this error captures ambiguity in robot communication.

\item \emph{Timing Failure.} The system introduces a deliberate delay of 3 secs (selected to disrupt the flow without causing total disengagement) before responding, followed by directing the human participant to the right drawer. This failure models disruptions in conversational timing and response latency.

\item \emph{Search Failure.} The robot identifies and highlights the wrong drawer through both speech and LED cues. This simulates  guidance errors that occur when robots mislocalize objects or misinterpret the task state, or even mishear the item mentioned by the participant. 

\item \emph{Comprehension Failure.} The robot signals that it cannot understand the participant’s request. This represents breakdowns in speech understanding, which are common in noisy or dynamic environments such as the fast-paced emergency rooms.
For example, after a participant requests a specific item (e.g., “epinephrine syringe”), the robot responds that it did not understand and asks for repetition, without offering any guidance.

\end{itemize}

\subsection{Study Procedure and Data Collection} \label{sec:study_procedure}

Each study session consisted of four phases: an introduction, a guided object–retrieval block, a post-study reflection survey, and an optional demographic and background survey.

\subsubsection{Introduction.}
After providing informed consent, participants were introduced to a crash cart robot designed to support item retrieval in a clinical setting. The facilitator explained that the robot’s role was to help users locate medical supplies efficiently by providing spoken instructions and visual cues, and that the participant’s task was to follow the robot’s direction to retrieve requested items as accurately and quickly as possible. The participants were not informed that the robot could behave incorrectly or that failures would be introduced.

\subsubsection{Guided Object–Retrieval Block.}
The core task consisted of a sequence of guided item-retrieval trials with the crash cart robot. Each participant completed a predetermined sequence of item retrieval trials with the crash cart robot. See Section \ref{sec:study_design} for details on the retrieval tasks and conditions. For each trial, the facilitator requested the participant to retrieve one item, and the robot responded with spoken instructions and visual cues to guide participants to the appropriate location.

\subsubsection{Post-Study Reflection.}

\begin{figure}[t]
\centering
\includegraphics[width=0.9\columnwidth]{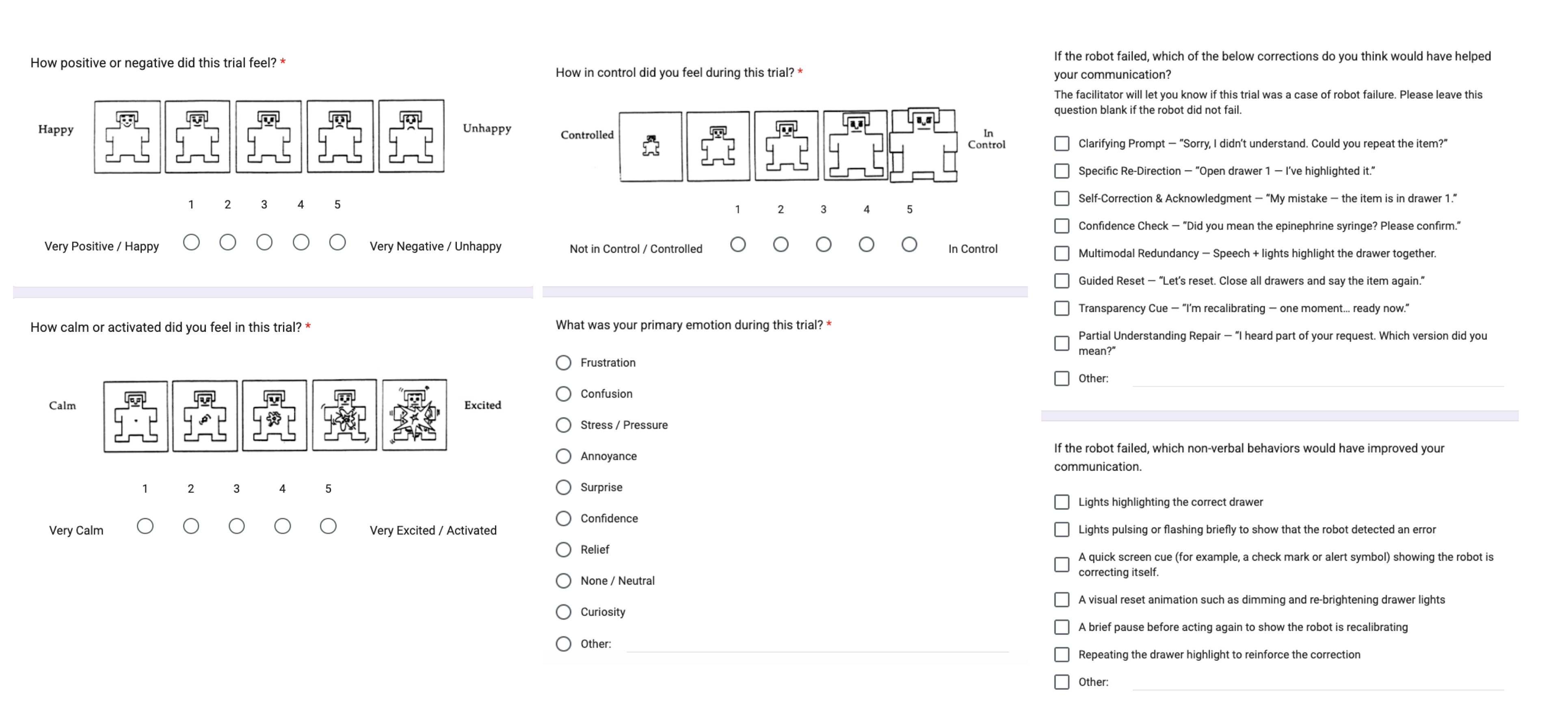}
\caption{Post-study survey form used to collect affective responses, including categorical emotion selection and Self-Assessment Manikin (SAM) ratings, as well as participants’ preferred recovery strategies following each trial.}
\label{fig:survey_details}
\end{figure}

We administered a survey to assess participants emotional reaction and recovery preferences. The full survey instruments are shown in Figure \ref{fig:survey_details}.

\begin{itemize}
\item \emph{Emotional reaction.} For each trial, participants provided two forms of affective self-report. First, they selected their dominant emotional reaction, choosing from frustration, confusion, stress/pressure, annoyance, surprise, confidence, relief, curiosity, neutrality, or an open-ended option. Second, they reported their affective state using the Self-Assessment Manikin (SAM), which measures valence, arousal, and perceived control~\cite{bradley1994measuring}.

\item \emph{Recovery preference.} Participants also indicated which of several candidate recovery strategies they would have preferred the robot to take for the observed trial (e.g., offering an explicit apology and clarification, providing more step-by-step guidance, or escalating to a human teammate).
\end{itemize}

\subsubsection{Demographic Survey.}
Participants were asked to complete a brief, optional demographic questionnaire to support descriptive analysis of the study sample. The survey collected basic background information, including age range, gender identity, prior experience interacting with robots or automated systems, and familiarity with clinical or hospital environments.
When time permitted, the demographic survey was administered immediately following the post-study reflection phase. In cases where session time was constrained, the same questionnaire was distributed to participants via follow-up email. All demographic responses were collected separately from interaction data and were used solely for aggregate analysis.

\section{Dataset}
\label{sec:Dataset}

\begin{figure}[t]
    \centering
    \begin{tikzpicture}[
        font=\small,
        node distance=7mm and 14mm,
        box/.style={draw, rounded corners=2pt, inner sep=4pt, align=left, anchor=west},
        group/.style={draw, rounded corners=2pt, inner sep=5pt, align=left},
        arrow/.style={-Latex, line width=0.55pt},
        pill/.style={draw, inner sep=3pt, align=center}
    ]


    \node[box, align=center, anchor=west, yshift=16mm, minimum width=35mm] (comp) {\textbf{Comprehension}\\ \textbf{Failure}\\ \footnotesize 42 trials};
    \node[box, align=center, anchor=west, below=4mm of comp.south, minimum width=35mm] (search) {\textbf{Search Failure}\\ \footnotesize 43 trials};
    \node[box, align=center, anchor=west, below=4mm of search.south, minimum width=35mm] (speech) {\textbf{Speech Failure}\\ \footnotesize 48 trials};
    \node[box, align=center, anchor=west, below=4mm of speech.south, minimum width=35mm] (timing) {\textbf{Timing Failure}\\ \footnotesize 40 trials};
    \node[box, align=center, anchor=west, below=4mm of timing.south, minimum width=35mm] (success) {\textbf{Success}\\ \footnotesize 41 trials};

    \node[
    group,
    fit=(comp)(success),
    inner sep=6pt,
    minimum width=35mm,
    label={[font=\normalsize\bfseries, yshift=15pt]above:Cleaned Dataset},
    label={[font=\small\bfseries, yshift=5pt]north:Trial Outcome Categories (5)}
    ] (outcomes) {};

    \node[
    group,
    inner sep=6pt,
    minimum width=35mm,
    align=center,
    yshift=10pt,
    right=18mm of outcomes](data){\textbf{Data Released}\\ \textbf{For Each Sample}};

    \coordinate (data-top)    at ($(data.north west)!0.15!(data.south west)$);
    \coordinate (data-upper)  at ($(data.north west)!0.35!(data.south west)$);
    \coordinate (data-middle) at ($(data.north west)!0.50!(data.south west)$);
    \coordinate (data-lower)  at ($(data.north west)!0.65!(data.south west)$);
    \coordinate (data-bottom) at ($(data.north west)!0.85!(data.south west)$);

    \draw[arrow] (comp.east)    -- (data-top);
    \draw[arrow] (search.east)  -- (data-upper);
    \draw[arrow] (speech.east)  -- (data-middle);
    \draw[arrow] (timing.east)  -- (data-lower);
    \draw[arrow] (success.east) -- (data-bottom);

    \node[anchor=west, align=center, right=15mm of data, yshift=40mm, minimum width=36mm, minimum height=5mm] (visualtitle)
    {\textbf{MediaPipe Folder}};
    
    \node[
        matrix,
        anchor=north west,
        inner sep=0pt,
        row sep=2mm,
        column sep=4mm,
        align=center,
    ] (visualpills) at ([yshift=-3mm]visualtitle.south west) {
        \node[pill, minimum width=35mm] (au) {\footnotesize au.csv \\ \footnotesize shape M(SD): (784(465), 18(0))}; \\
        \node[pill, minimum width=35mm] (gaze) {\footnotesize gaze.csv \\ \footnotesize shape M(SD): (784(465), 7(0)}; \\
        \node[pill, minimum width=35mm] (pose) {\footnotesize pose.csv \\ \footnotesize shape M(SD): (784(465), 7(0)}; \\
        \node[pill, minimum width=35mm] (landmarks) {\footnotesize landmarks.csv \\ \footnotesize shape M(SD): (784(465), 1435(0)}; \\
    };
    
    \node[
        group,
        fit=(visualtitle)(visualpills),
        inner sep=6pt
    ] (visual) {};

    \node[box, right=13mm of data, yshift=-15mm, align=center, minimum width=40mm, minimum height=10mm] (subjective) {\textbf{Survey Data}\\ \footnotesize survey.csv \\ shape M(SD): (1(0), 6(0))};
    \node[box, right=13mm of data, yshift=-30mm, align=center, minimum width=40mm, minimum height=10mm] (language) {\textbf{Speech Data}\\ \footnotesize transcript.txt \\ \footnotesize length M(SD): 411.8 (222.6) chars};
    \node[box, right=13mm of data, yshift=-55mm, align=center, minimum width=40mm, minimum height=10mm] (language) {\textbf{Metadata}\\ \footnotesize meta.json \\ \footnotesize trial start and end timestamps, \\ \footnotesize trial duration, trial order, \\ \footnotesize participant ID, sample
ID, participant \\ \footnotesize  status: healthcare expert \\ \footnotesize or non-expert};

    \draw[arrow] (data.east)  -- (visual.west);
    \draw[arrow] (data.east)  -- (subjective.west);
    \draw[arrow] (data.east)  -- (language.west);

    \end{tikzpicture}
    \caption{\textbf{Dataset organization and per-trial multimodal composition.} 
The cleaned dataset is organized into five trial outcome categories (four injected robot failure types and success). For each trial, we release a self-contained sample comprising time-indexed visual behavioral features (MediaPipe-derived facial action units, gaze, pose, and landmarks), structured post-trial survey responses, and a natural-language task transcript. An overview of the dataset structure and accessible modalities is illustrated in Fig.~\ref{fig:dataset_overview}. Together, these views summarize both the dataset-level structure and the multimodal signals available for modeling trial outcomes and recovery preferences, while abstracting away file-level organization.}
    \label{fig:dataset_overview}
\end{figure}
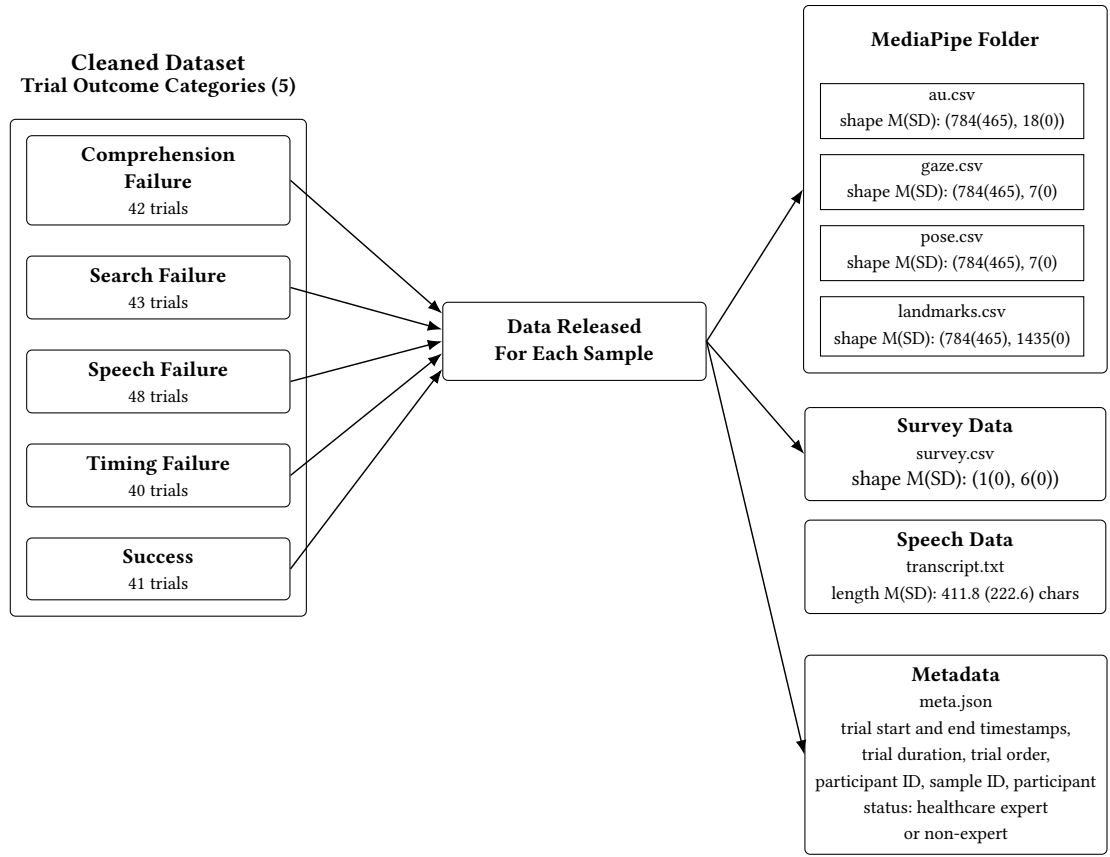

\subsection{Annotation Protocol and Labels}

\subsubsection{System-Level Failure and Trial Boundary Annotation}
\label{sec:trial_boundaries}
System-level annotations capture both (i) trial boundaries and (ii) robot communication failure types. Labels were assigned by a team of five trained researchers through manual review of recorded session videos, informed by the robot’s internal logs and observable robot outputs. Each trial was annotated by a single researcher.

Trial segmentation was performed using consistent criteria across sessions. Trial start was defined as the onset of the facilitator’s verbal request for the next item. Trial end was defined as the final spoken utterance before the subsequent trial began, or the end of the recording if no subsequent trial occurred.

Failure annotations indicate whether the robot behaved as designed or exhibited one of four scripted communication failures: speech, timing, search, or comprehension. Each failure type was associated with characteristic robot behaviors and utterances that served as cues during annotation. Speech failures involved under-specified instructions (e.g., “Open the drawer”) or the absence of speech; timing failures involved delayed responses with apologetic cues (e.g., “Sorry, I am delayed…”); search failures involved incorrect spatial guidance (e.g., “I think the item is in drawer \_\_”); and comprehension failures involved explicit non-understanding (e.g., “I did not understand that request.”). System logs, verbal outputs, and visual context were jointly used to disambiguate failure types.

\subsubsection{Annotation Workflow and Consistency}
Annotations were produced by trained members of the research team following a shared, written annotation protocol that defined trial boundaries and failure categories (see Section \ref{sec:trial_boundaries}). Annotators reviewed recorded session videos and referenced both system logs and observable robot behavior when assigning labels. Consistency was supported through the use of common definitions for trial start and end, explicit mappings between robot behaviors and failure types, and discussion-based adjudication of ambiguous cases.
Participant-provided annotations were collected separately during the post-study reflection phase, as described below.

\subsection{Data Processing and Feature Extraction}
In our annotations, we focus \emph{system errors}; we do not annotate or release user-error labels. 
We transformed the raw recordings into a set of aligned multimodal feature streams suitable for training and benchmarking machine learning algorithms. At a high level, the dataset includes four types of signals per trial. These include structured trial-level metadata, facial and head pose features, transcribed speech features, and survey responses.
Detailed descriptions of these signals, and how they were extracted, can be found in the sections below.

\subsubsection{Trial Segmentation and Metadata} Each participant session was segmented into trial-level interactions, where each trial corresponds to a single item retrieval attempt from initiation to completion. Trial segmentation was performed manually by trained researchers through review of recorded session videos, and clear definitions of what constituted trial start and trial end (see Section \ref{sec:trial_boundaries}).
Using these boundaries, we recorded structured trial-level metadata including failure type, trial start and end timestamps, participant identifier, trial index, sample identifier, and whether the participant was a healthcare expert or non-expert.

\subsubsection{Facial and Head Pose Features} \label{sec:facial_head_pose_features}
We analyze frame-level visual descriptors captured from the egocentric video, reflecting participants’ facial behavior, head orientation, and visual attention during interaction with the crash cart robot. Specifically, we extract all visual features using Google MediaPipe’s face analysis pipelines ~\cite{lugaresi2019mediapipe}, which provide robust real-time estimation of facial landmarks and head pose from monocular RGB video.

For each session, we process the video stream with MediaPipe Face Mesh; this tool detects a dense set of 468 3D facial landmarks per frame, which use normalized image-space coordinates and relative depth. These landmarks encode fine-grained facial geometry across regions such as the eyebrows, eyes, nose, lips, and jawline. Using the detected landmarks, we estimate a 3D rigid head pose for the participant at each frame, defining it by rotation angles (pitch, yaw, roll) and translation with respect to the camera. We additionally derive coarse gaze-related features from eye-region landmarks, capturing attention shifts and visual search behavior during interaction breakdowns.
In parallel, we extract frame-level facial action unit (AU) intensities using MediaPipe’s Face Landmarker model with blendshape outputs. We map relevant blendshape coefficients to a subset of OpenFace-style action units \cite{baltrusaitis2018openface} and scale their magnitudes to produce continuous AU intensity estimates per frame.

For downstream modeling and dataset release, we retain only frames with successful face detection and store the resulting facial landmark, head-pose, gaze-related, and facial action unit features as synchronized per-frame time series indexed by video frame number.
Due to the distributed nature of the data collection effort, we did not enforce standardized cameras or video resolutions across recording sessions. Instead, we standardized the extracted behavioral signals during post-processing by operating on normalized facial landmarks and by producing temporally aligned and intensity-normalized facial action unit features, enabling consistent downstream analysis across heterogeneous video inputs.

\subsubsection{Transcribed Speech Features} 
To capture speech data while preserving participant privacy, we converted the audio data of each session into time-stamped transcripts using the Whisper Speech Recognition model~\cite{radford2022whisper}. Whisper utilizes a Transformer-based encoder–decoder architecture, which researchers trained on hundreds of thousands of hours of multilingual speech. We use the model in English transcription mode to obtain segment-level transcripts with start and end timestamps.
We retain the resulting time-stamped text segments for downstream analysis of participant–robot interaction dynamics.
To balance privacy with the ability to model temporal dynamics in user reactions to robot failures, we release derived text features and time-stamped transcripts, but not the raw audio recordings.

\subsubsection{Survey-Based Behavioral Signals}
\label{sec:survey_features}

We include structured survey responses capturing participants’ subjective experience of robot behavior on a per-trial basis, collected during the post-study reflection phase (see Section~\ref{sec:study_procedure}). For each trial, we release two categories of survey-derived signals.
Participants’ \emph{emotional reactions} are represented using a categorical label indicating the dominant reported emotion and continuous affective ratings along the valence, arousal, and perceived control dimensions obtained using the SAM scale~\cite{bradley1994measuring}.
Participants’ \emph{recovery preferences} are represented as categorical labels corresponding to the single recovery strategy selected for each trial.
All survey responses are released as structured, trial-level annotations aligned with the corresponding trial metadata.
The full survey instruments are shown in Figure \ref{fig:survey_details}.

\section{Data Analysis}
We analyzed the dataset to  characterize participant reactions to robot failures during object retrieval, and  quantify how self-reported affect, discrete emotions, and recovery preferences vary across failure types, and success vs. failure outcomes. 

\subsection{Dataset Exploration and Quality Control}
We performed a dataset-wide exploratory audit to characterize composition, balance, and the repeated-measures structure prior to inferential testing. We computed condition-wise distributions of  the number of participants and samples, including expert/non-expert breakdowns, the frequency distribution of reported emotions, and recovery preference selections stratified by modality (verbal, nonverbal, multimodal, and no recovery needed), providing a descriptive baseline for subsequent analyses.

We examined all (N=214) samples to assess the availability and integrity of the expected multimodal components (metadata, survey responses, speech transcripts, and visual behavioral features). For each modality, we checked schema-level consistency (e.g., expected fields/columns and data dimensional structure), identified and resolved issues such as missing or malformed data, and harmonizing representations where needed, and subsequently verified that the resulting data supported comparable analyses.

To characterize the repeated-measures structure, we quantified the samples-per-participant distribution and verified participant-to-sample mappings, to document the non-independent observation structure, which guided selection of statistical methods accounting for within-participant correlation in subsequent analyses. 
These checks were also used to confirm deviations from the nominal study design: 3 participants experienced fewer than 5 conditions and 4 participants experienced more than five conditions. Because participants reflected on these as unified experiences in their post-study responses, we excluded  samples with multiple labels (conditions) from condition-based analyses to avoid ambiguous condition assignment, while retaining them in the released dataset. 

We also quantified variation in survey instrument versions: samples from earlier participants (samples 1–102) included only verbal recovery strategy options, whereas later participants (samples 103–214) received an expanded survey with verbal, nonverbal, and multimodal options of recovery preferences. This variation is accounted for in our recovery preference analyses by reporting results separately by survey version and avoiding direct comparisons between strategies that appeared on different instruments.
All subsequent analyses were conducted on the dataset after integrity verification and harmonization, with residual inconsistencies handled via predefined exclusions or carried forward as analysis constraints.

\subsection{Primary Emotion Analysis}
\label{subsec:primary-emotion-analysis}

We analyzed self-reported \textit{primary emotions} to characterize affective responses to robot performance during object retrieval.  Only trials with a single  condition label were included in inferential analyses. The analysis addressed four questions: (i) how primary emotions are distributed overall, (ii) whether emotion distributions differ by outcome (Success vs.\ Failure), (iii) whether they differ across condition types, and (iv) whether emotions change over time (trial order).

\subsubsection{Descriptive Characterization.}
We first computed overall frequencies and percentages of primary emotions across all trials, followed by stratified summaries by outcome and by condition type. This step provided an interpretable baseline, identified dominant and sparse emotion categories, and guided which emotions were suitable for downstream modeling.

\subsubsection{Emotions by Outcome and Conditions.}
We performed inspection of the stratified distributions that revealed clear variation in emotion prevalence across both outcome and condition types. To evaluate whether these observed differences reflected systematic effects rather than random variation, we tested for \textbf{associations} between primary emotion and (i) outcome (Success vs.\ Failure) and (ii) condition type (five levels: four failures and success). In both analyses, the independent variable (IV) was categorical (outcome or condition), and the dependent variable (DV) was the categorical primary-emotion label. The contingency-based analyses assume categorical variables and independent observations. While the measurement assumptions are satisfied by design, independence is violated due to repeated measures within participants. We therefore interpret chi-square statistics as global measures of association and rely on within-participant permutation tests for valid significance estimation. Association strength is quantified using Cramér’s $V$, which is emphasized over significance alone to characterize the magnitude of distributional differences.

\subsubsection{Accounting for Repeated Measures.}
Because each participant contributed to multiple trials, trial-level observations were not independent. To obtain valid significance estimates under this repeated-measures structure, we computed permutation-based $p$-values by shuffling outcome or condition labels \emph{within participants}. This preserves individual emotion response profiles while testing whether the alignment between task conditions and emotions exceeds chance expectations under clustering.

\subsubsection{Temporal Analysis}
To examine whether emotional responses evolved over time, we analyzed emotion patterns as a function of trial index. Trial-wise emotion frequencies were first inspected descriptively. For inferential quantification, we modeled the most frequent emotions as binary outcomes (presence vs.\ absence of a target emotion) using participant-clustered logistic models. Specifically, we estimated population-average effects with \textbf{generalized estimating equations (GEE)}, using trial index and outcome as predictors and clustering on participant. GEE provides robust inference under misspecification of the working correlation, making it well-suited for the present repeated-measures, sparse-category setting.
 GEE was selected because likelihood-based mixed-effects logistic models were unstable in this dataset due to sparse emotion categories and quasi-separation, leading to convergence failures and unreliable variance estimates. Emotions with insufficient occurrences were excluded from emotion-specific models.

\paragraph{Summary.}
Overall, descriptive analyses characterize the emotional landscape of interaction outcomes, contingency-based association tests validate global distributional differences across outcome and condition type, and participant-aware logistic models quantify how specific emotions vary with task progression over time.

\subsection{Affective responses analysis}

The affective analysis addressed three primary objectives:
\begin{enumerate}
    \item To determine whether affective responses differ between successful and failed interactions.
    \item To assess whether affective responses vary across the five interaction conditions (Success and four failure types).
    \item To evaluate the potential influence of trial order on affective responses.
\end{enumerate}

\subsubsection{Data Structure}
Affective responses consisted of three 5-point ordinal variables (valence, arousal, and perceived control), recorded once per sample. We analyzed a subset restricted to one condition per sample (see Section 5.1). Given the ordinal, bounded nature of SAM ratings and the presence of within-participant dependence, all inferential procedures were selected to avoid assumptions of normality and sample-level independence.

\subsubsection{Descriptive Characterization}
As an initial step, we computed descriptive summaries for valence, arousal, and perceived control across the full dataset and stratified by interaction outcome and condition. Medians and interquartile ranges were used as primary descriptors, supplemented by means and standard deviations for completeness. This descriptive analysis provided an overview of affective response patterns and informed subsequent inferential test selection.

\subsubsection{Affective Responses by Outcome}
Examination of outcome-stratified affective rating distributions revealed clear and consistent shifts between successful and failed interactions. This strong descriptive structure motivated formal testing of whether these differences represent systematic within-participant effects.
For inferential analysis, each affective dimension was analyzed separately. The independent variable was interaction outcome (Success vs. Failure), and the dependent variable was the corresponding affective rating.

To respect the repeated-measures structure (several samples/trials per participant), affective ratings were aggregated at the participant level. For each participant, the median affective rating across success samples and the median affective rating across failure samples were computed, yielding one paired observation per participant.
Paired differences were evaluated using the Wilcoxon signed-rank test, a non-parametric test that does not assume normality. Effect sizes were quantified using the matched-pairs rank-biserial correlation, providing an interpretable measure of the magnitude and direction of within-participant differences.

\subsubsection{Affective Responses by Conditions.}
We examined condition-stratified distributions of affective ratings to characterize how valence, arousal, and perceived control vary across the five interaction conditions. These descriptive patterns were subsequently assessed using participant-level inferential analyses to evaluate condition-related differences in affective responses within participants.
Each affective dimension (dependent variable) was analyzed separately, with interaction condition as a five-level categorical independent variable. 
Because each participant experienced multiple conditions, analyses were conducted within a repeated-measures (within-participant) framework. Condition-level inference was restricted to participants who contributed at least one valid observation for all five conditions. For participants with multiple trials within a condition, the median affective rating for that condition was used.
Differences across conditions were assessed using the Friedman test (the non-parametric analogue of repeated-measures ANOVA). We quantified effect sizes using Kendall’s $W$, capturing the degree of systematic variation in affective responses across conditions at the participant level.

\subsubsection{Handling of Repeated Measures and Model Assumptions}
Across all analyses, repeated measures (multiplicity of samples/trials per participant) were handled either by aggregating affective ratings within participant prior to testing or by using explicitly repeated-measures non-parametric tests. We avoided trial-level tests assuming independent observations. Linear mixed-effects models were explored as a sensitivity analysis but were not retained for inference due to identifiability issues arising from the limited and unbalanced condition--trial structure. Accordingly, rank-based within-participant methods were adopted as the primary inferential approach.

\subsubsection{Temporal Considerations}
Because trials were completed sequentially, affective responses may also reflect time-on-task or interaction fatigue effects independent of condition. To examine this possibility, we analyzed affective ratings descriptively as a function of trial index. Trial order was not treated as a primary inferential factor but was used to contextualize condition effects, as descriptive inspection did not reveal consistent or observable changes in affective ratings across trials.

\paragraph{Summary}

In summary, we analyzed affective responses   using a participant-aware, non-parametric framework that respects the ordinal nature of SAM ratings, accounts for \textbf{within-subject dependence}, and separates analytical procedures from reported results. All statistical outcomes and condition-specific patterns are presented in Section~\ref{sec:affect_results}

\subsection{Recovery Strategy Analysis}

We analyzed participants’ reported recovery strategy selections to characterize the recovery design space represented in the dataset and to document how recovery behaviors were distributed across communication modalities and strategy types. This analysis is descriptive in nature and focuses on dataset composition and structure.  While modality-level summaries were computed on restricted, comparable subsets (see Section 5.1), recovery strategy selections overall varied in number across participants; accordingly, we do not draw inferential claims about recovery preferences or optimal strategies at this stage.



\subsubsection{Strategy Identification and Categorization.}
Recovery responses were parsed by separating comma-delimited selections into individual strategy mentions and normalizing textual variants. Across the dataset, this process yielded a set of distinct recovery strategies, including predefined survey options and participant-provided write-in responses, resulting in a total of 34 unique recovery strategy types, including strategies derived from expanded free-text responses. These strategies were retained as distinct labels for counting and reporting purposes.
Each strategy was further categorized along two complementary dimensions. First, strategies were grouped by communication modality into \emph{verbal}, \emph{nonverbal}, and \emph{multimodal} categories, where multimodal strategies explicitly combined speech with visual or embodied cues (e.g., simultaneous verbal explanation and visual highlighting). Second, verbal strategies were grouped by communicative function (e.g., clarification request, acknowledgment of error, explicit redirection, transparency cue).

\subsubsection{Descriptive Summaries and Normalization.}
We computed frequency counts of recovery strategy mentions at multiple levels of aggregation, including per-strategy counts, per-category counts, and per-modality counts. Because participants could select multiple recovery strategies within a single trial, all summaries are reported in terms of mentions rather than trials. For modality-level comparisons, counts were normalized within the eligible subset of failure trials to report proportions that reflect relative composition rather than absolute sample size.

\section{Results}

In this section, we present preliminary results from an exploratory analysis of the RFM-HRI dataset. These analyses are intended to illustrate the range of affective, behavioral, and recovery-related signals captured in the data, rather than to draw causal conclusions.
We summarized the dataset’s composition and then examined user responses to robot failures during crash cart item retrieval, focusing on affective reactions across failure types and interaction outcomes. Finally, we reported initial patterns in user preferences for robot recovery behaviors, highlighting  signals for failure acknowledgment and guidance.
Together, these exploratory results operationalize the core contributions of this work: a multimodal, healthcare-inspired HRI dataset, an empirical characterization of user responses to robot failures, and early design-relevant insights into preferred recovery strategies. While preliminary, these findings underscore the utility of RFM-HRI for research on failure detection, user state modeling, and recovery policy design in healthcare and broader workplace robotics contexts.

\subsection{Descriptive Statistics}

The dataset contains \textbf{214 samples}: 64 of which were collected in the hospital with healthcare workers and 136 samples in laboratory settings with non-experts. 
For each sample, metadata files contain 6 keys capturing trial- and participant-level identifiers. Survey data consist of 6 questions per trial and cover \textbf{13 distinct emotion categories in total} and \textbf{34 unique recovery preference types} across all conditions. Visual features from image frame counts per trial had a mean of 784 (SD = 465). Verbal interaction transcripts had an average length of 411.8 characters per trial (SD = 222.6). An overview of dataset composition across conditions is provided in Table~\ref{tab:dataset_overview}.

\begin{table*}[t]
\centering
\footnotesize
\setlength{\tabcolsep}{4pt}
\caption{Dataset overview across entire dataset (214 samples)} 
\label{tab:dataset_condition_summary}
\begin{tabularx}{\textwidth}{l l c X X}
\toprule
\textbf{Condition} &
\textbf{Participants} &
\textbf{\# Samples} &
\textbf{All emotions (count)} &
\textbf{Recovery preferences (count)} \\
\midrule

Comprehension &
38 Total (11 Expert,  27 Non-Expert) &
42 (13 Expert, 29 Non-Expert)&
\begin{tabular}[t]{@{}l@{}}
Annoyance (11) \\
Confusion (11) \\
Frustration (10) \\
Curiosity (4) \\
Surprise (3) \\
Confidence (2) \\
Relief (1)
\end{tabular}
&
\begin{tabular}[t]{@{}l@{}}
Verbal (111) \\
Nonverbal (1) \\
Multimodal (17) \\
No recovery needed (0)
\end{tabular}
\\
\addlinespace

Search &
39 Total (12 Expert,  27 Non-Expert) &
43 (15 Expert, 28 Non-Expert) &
\begin{tabular}[t]{@{}l@{}}
Confusion (16) \\
Annoyance (14) \\
Frustration (5) \\
Stress / pressure (3) \\
Confidence (2) \\
Curiosity (2) \\
None / neutral (1)
\end{tabular}
&
\begin{tabular}[t]{@{}l@{}}
Verbal (110) \\
Nonverbal (2) \\
Multimodal (17) \\
No recovery needed (0)
\end{tabular}
\\
\addlinespace

Speech &
40 Total (11 Expert,  29 Non-Expert) &
48 (15 Expert, 33 Non-Expert)&
\begin{tabular}[t]{@{}l@{}}
Confusion (17) \\
Annoyance (9) \\
Frustration (6) \\
Confidence (5) \\
Surprise (4) \\
Relief (2) \\
Stress / pressure (2) \\
Amused (1) \\
Curiosity (1) \\
None / neutral (1)
\end{tabular}
&
\begin{tabular}[t]{@{}l@{}}
Verbal (104) \\
Nonverbal (5) \\
Multimodal (36) \\
No recovery needed (1)
\end{tabular}
\\
\addlinespace

Success &
37 Total (10 Expert,  27 Non-Expert) &
41 (13 Expert, 28 Non-Expert)&
\begin{tabular}[t]{@{}l@{}}
Surprise (12) \\
Relief (10) \\
Confidence (6) \\
Confusion (4) \\
Annoyance (3) \\
Frustration (2) \\
Curiosity (1) \\
Happy (1) \\
In doubt (1) \\
None / neutral (1)
\end{tabular}
&
\begin{tabular}[t]{@{}l@{}}
Verbal (23) \\
Nonverbal (0) \\
Multimodal (7) \\
No recovery needed (2)
\end{tabular}
\\
\addlinespace

Timing &
36 Total (10 Expert,  26 Non-Expert) &
40 (12 Expert, 28 Non-Expert)&
\begin{tabular}[t]{@{}l@{}}
Frustration (12) \\
Confusion (11) \\
Annoyance (10) \\
Curiosity (2) \\
Surprise (2) \\
Interest (1) \\
Confidence (1) \\
Relief (1)
\end{tabular}
&
\begin{tabular}[t]{@{}l@{}}
Verbal (113) \\
Nonverbal (2) \\
Multimodal (13) \\
No recovery needed (0)
\end{tabular}
\\

\bottomrule
\end{tabularx}
\label{tab:dataset_overview}
\end{table*}

\subsection{Primary Emotions}
\label{sec:results_primary_emotions}

\subsubsection{Analysis Set}  From 214 total samples, 25 trials contained multiple condition labels (e.g., trials where two failures occurred consecutively) and were excluded from condition-based inferential tests. The final analysis set comprised $N=189$ trials from 41 participants (trials per participant: 1--5, $M=4.6$). Of these, 153 were failure trials and 36 were success trials.

\subsubsection{Descriptive Characterization.} Across all analyzed trials ($N=189$), the most frequent emotions were Confusion ($N=49$, 25.9\%), Annoyance ($N=42$, 22.2\%), and Frustration ($N=33$, 17.5\%), followed by Surprise ($N=19$, 10.1\%), Confidence ($N=16$, 8.5\%), Relief ($N=12$, 6.3\%), Curiosity ($N=8$, 4.2\%), and Stress/Pressure ($N=5$, 2.6\%). Five categories were sparse ($N<5$ each: Amused, None/Neutral, In Doubt, Happy, Interest) (Table \ref{tab:overall_emotion_distribution}).

\begin{table}
\centering
\small
\caption{Overall distribution of primary emotions across  analyzed samples ($N=189$).}
\label{tab:overall_emotion_distribution}
\setlength{\tabcolsep}{6pt}
\begin{tabular}{lrr}
\toprule
\textbf{Emotion} & \textbf{Count} & \textbf{Percentage (\%)} \\
\midrule
Confusion            & 49 & 25.9 \\
Annoyance            & 42 & 22.2 \\
Frustration          & 33 & 17.5 \\
Surprise             & 19 & 10.1 \\
Confidence           & 16 & 8.5 \\
Relief               & 12 & 6.3 \\
Curiosity            & 8  & 4.2 \\
Stress / Pressure    & 5  & 2.6 \\
Amused               & 1  & 0.5 \\
None / Neutral       & 1  & 0.5 \\
In Doubt             & 1  & 0.5 \\
Happy                & 1  & 0.5 \\
Interest             & 1  & 0.5 \\
\midrule
\textbf{Total}       & \textbf{189} & \textbf{100.0} \\
\bottomrule
\end{tabular}
\end{table}

\subsubsection{Emotions by Outcome.} 
\begin{figure}[t]
    \centering
    \includegraphics[width=\columnwidth]{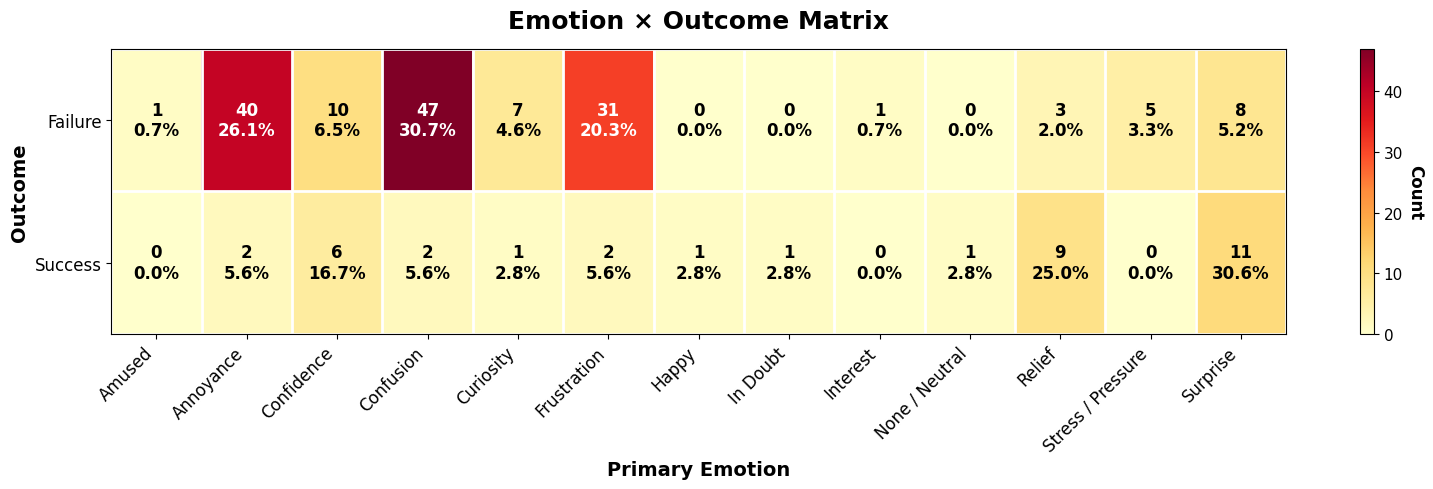}
    \caption{Primary emotion distributions by outcome (Success vs. Failure)}
    \label{fig:emotionxoutcm}
\end{figure}

\begin{figure}[t]
    \centering
    \includegraphics[width=\columnwidth]{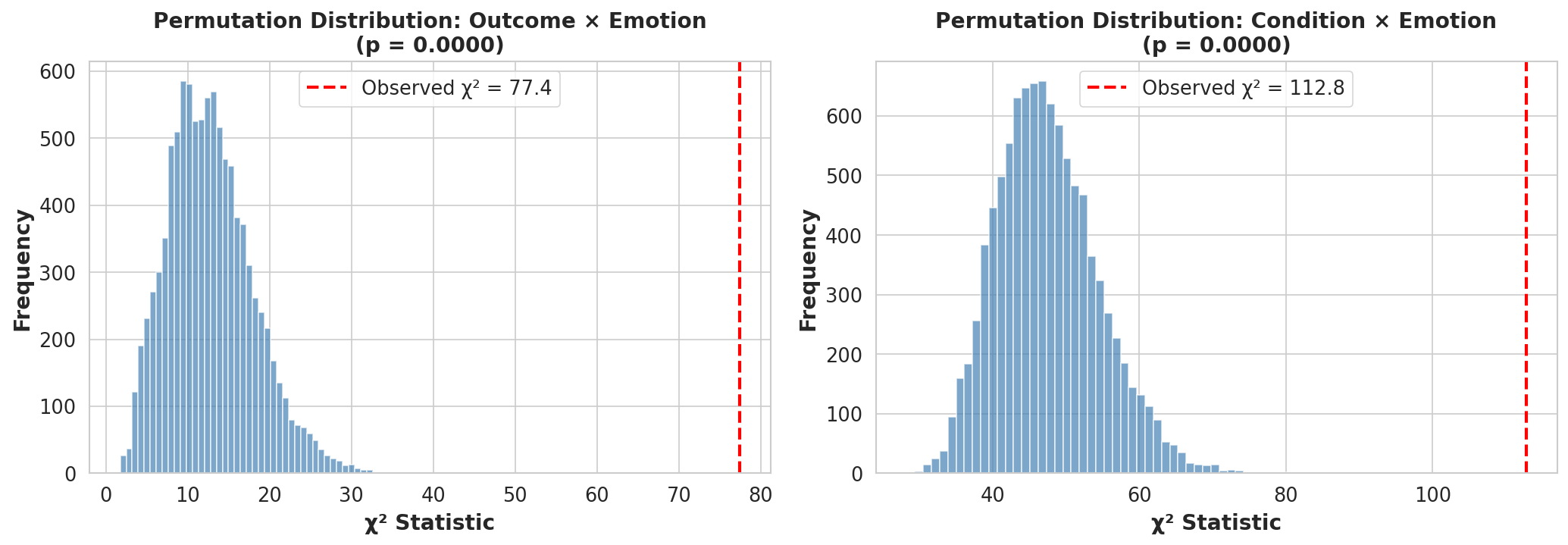}
    \caption{Permutation distributions for emotion -- outcome (left) and  emotion -- condition (right) associations.}
    \label{fig:emotion_sts}
\end{figure}

Emotion distributions differed substantially by outcome (Success vs.\ Failure). Failure trials were dominated by negative cognitive-evaluative states: Confusion ($47/153$, 30.7\%), Annoyance ($40/153$, 26.1\%), and Frustration ($31/153$, 19.1\%). In contrast, success trials most often elicited Surprise ($11/36$, 30.6\%), Relief ($9/36$, 25.0\%), and Confidence ($6/36$, 16.7\%) (Figure~\ref{fig:emotionxoutcm}). A chi-square test on the full $2\times 13$ contingency table showed a significant association between outcome and emotion, $\chi^2(12)=77.40$, $p<.001$, with a large effect size (Cramér’s $V=0.640$). Because trials are clustered within participants, we additionally computed participant-aware permutation $p$-values by shuffling outcome labels \emph{within} each participant (10{,}000 permutations), yielding a permutation $p<.0001$ (reported as $p=0.0000$ with 10{,}000 permutations). Figure~\ref{fig:emotion_sts} visualizes these outcome-level differences.

\subsubsection{Emotion by Condition.} Emotion distributions also varied across the five conditions (Success, Speech Failure, Timing Failure, Comprehension Failure, Search Failure). The $5\times 13$ chi-square test indicated a significant association between condition type and emotion, $\chi^2(48)=112.76$, $p<.001$, with a medium effect size (Cramér’s $V=0.386$). A within-participant permutation test (10{,}000 permutations; shuffling condition labels within participant) again yielded $p<.0001$. These confirm that the observed association was unlikely to arise from participant-specific response patterns or violations of sample independence.

\begin{figure}
    \centering
    \includegraphics[width=\columnwidth]{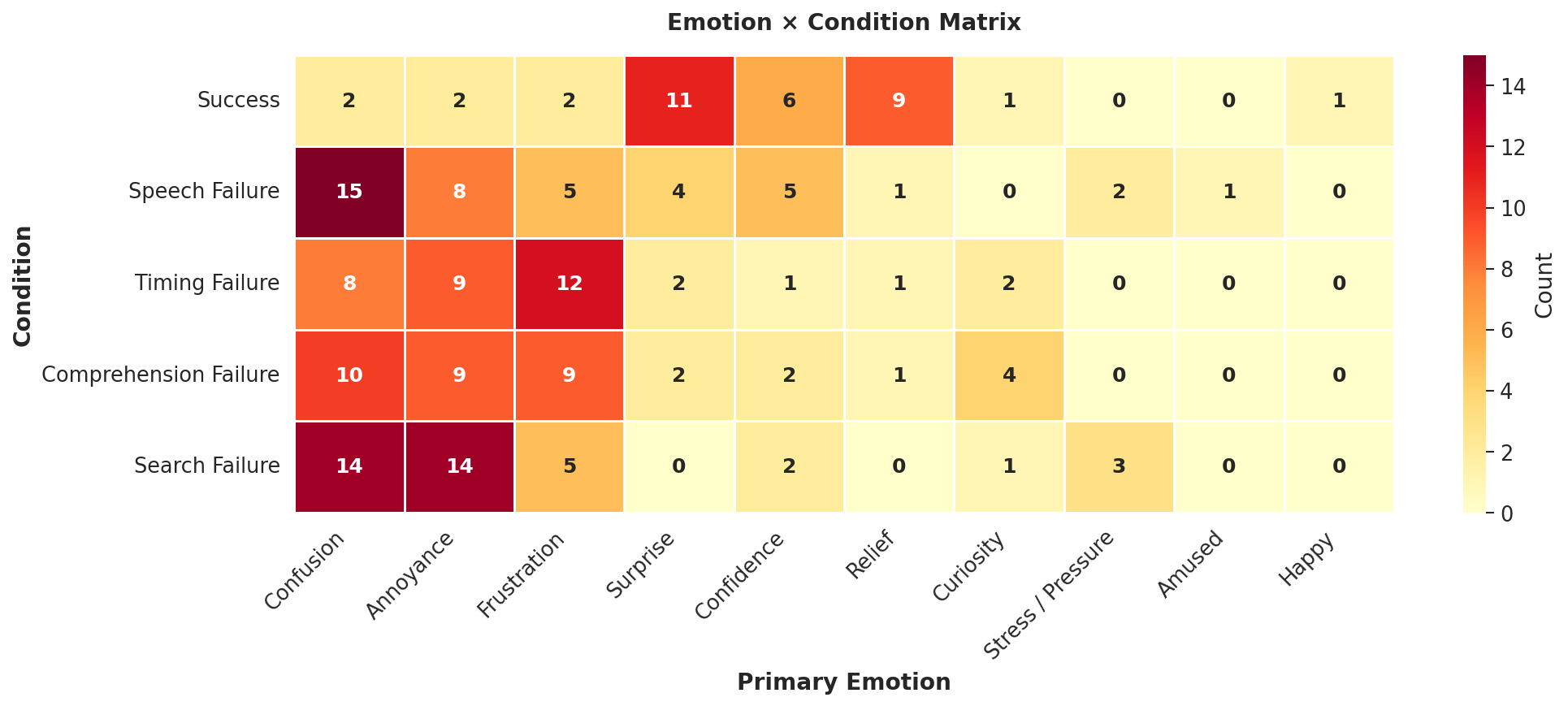}
    \caption{Primary emotion distributions across interaction conditions}
    \label{fig:emotionxcnd}
\end{figure}

Condition-wise counts are shown in Figure~\ref{fig:emotionxcnd}. Confusion was most frequent for Speech Failure ($N=15$) and for Search Failure ($N=14$), while Frustration was especially prominent for Timing Failure ($N=12$). Success trials were characterized by high Surprise ($N=11$) and Relief ($N=9$), with comparatively low Confusion ($N=2$) and Annoyance ($N=2$).

\paragraph {Temporal Analysis.} 

\begin{figure}[t]
    \centering
    \includegraphics[width=\columnwidth]{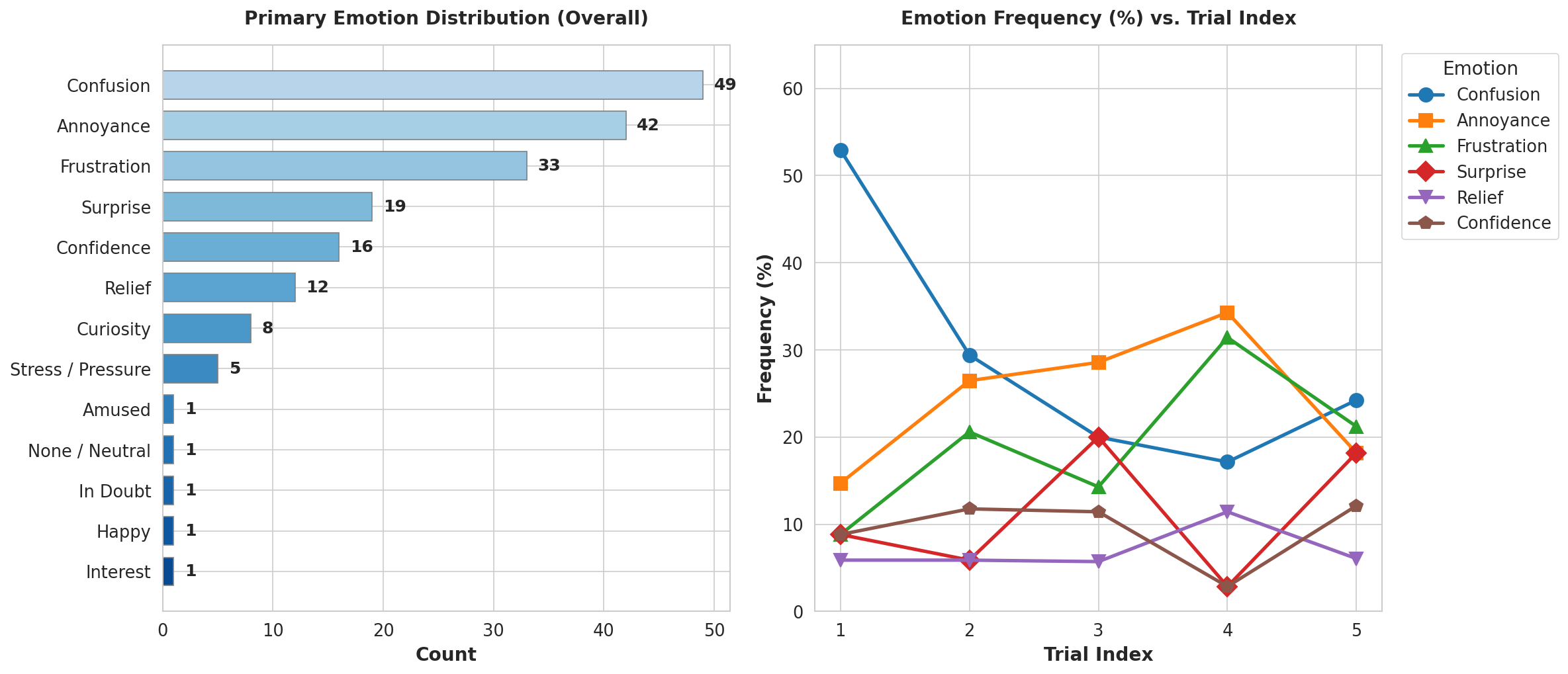}
    \caption{Overall and temporal distributions of primary emotions}
    \label{fig:emotion_allcnd}
\end{figure}

Trial-wise emotion frequencies revealed systematic changes across the five trial indices (1–5). Confusion was most prevalent early in the session (45.0\% in Trial 1) and declined across subsequent trials, whereas Frustration increased over time, peaking in Trial 4 (28.2\%). Annoyance showed a moderate increase through Trial 4 (30.8\%) followed by a decrease in the final trial (17.1\%). In contrast, emotions associated with resolution or positive appraisal (e.g., Surprise, Relief, Confidence) exhibited relatively stable frequencies across trials without pronounced monotonic trends (Figure \ref{fig:emotion_allcnd}).
Participant-clustered GEE logistic models confirmed these temporal patterns for the most prevalent emotions: trial index was associated with a significant decrease in the odds of reporting Confusion ($\beta = -0.289, p = 0.021$) and a significant increase in the odds of reporting Frustration ($\beta = 0.288, p = 0.017$), while the effect for Annoyance was not significant ($\beta = 0.118, p = 0.31$).

\subsection{Affective Response Analysis (SAM Ratings)}
\label{sec:affect_results}

\paragraph{Analysis Set}
Participants reported affective responses after each trial using the Self-Assessment Manikin (SAM), providing ratings along three dimensions: \emph{Valence} (unpleasant--pleasant), \emph{Arousal} (calm--excited), and \emph{Control} (low--high sense of control). Ratings were collected on 5-point ordinal scales. 
The affect dataset comprised 189 trials from 41 participants. Twenty-five trials with multiple condition labels were excluded to ensure unambiguous condition attribution. No trials were excluded due to missing or invalid affect data. 

\paragraph{Descriptive Characterization.}
Across all trials, median affective ratings indicated moderately negative experiences overall (Table~\ref{tab:affect_overall}). Valence ratings were centered toward the unpleasant end of the scale, while arousal and control ratings were moderate, suggesting emotionally engaged but not highly activated interactions.

\begin{table}[t]
\centering
\small
\caption{Overall affective ratings across all trials ($N=189$).}
\label{tab:affect_overall}
\begin{tabular}{lcccc}
\toprule
\textbf{Dimension} & \textbf{Median} & \textbf{IQR} & \textbf{Mean} & \textbf{SD} \\
\midrule
Valence  & 4.0 & [2.0--4.0] & 3.28 & 1.38 \\
Arousal  & 3.0 & [2.0--4.0] & 2.92 & 1.18 \\
Control  & 3.0 & [2.0--4.0] & 2.83 & 1.33 \\
\bottomrule
\end{tabular}
\end{table}

\paragraph{Affective Responses by Outcome}
we conducted analysis to 34 participants with data in both outcome categories (Success, Failure).
Paired Wilcoxon signed-rank tests revealed robust outcome-level differences across all three affective dimensions (Table~\ref{tab:affect_outcome}). Successful interactions were associated with significantly more positive valence (median$_{\text{Succ}}=1.0$ vs.\ median$_{\text{Fail}}=3.8$, $W=1.0$, $p=1.73\times10^{-6}$, $r_{\mathrm{rb}}=0.996$), lower arousal (median$_{\text{Succ}}=2.0$ vs.\ median$_{\text{Fail}}=3.0$, $W=81.0$, $p=0.0158$, $r_{\mathrm{rb}}=0.538$), and higher perceived control (median$_{\text{Succ}}=4.0$ vs.\ median$_{\text{Fail}}=2.5$, $W=88.0$, $p=0.00165$, $r_{\mathrm{rb}}=-0.645$) compared to failed interactions. Effect sizes indicated large effects for valence and control and a moderate effect for arousal.

\begin{table}[t]
\centering
\small
\caption{Paired Wilcoxon comparisons of affective ratings between Success and Failure outcomes (participant-level summaries, $n=34$).}
\label{tab:affect_outcome}
\begin{tabular}{lcccc}
\toprule
\textbf{Dimension} & \textbf{W} & \textbf{p-value} & \textbf{Rank-biserial $r$} & \textbf{Median (Succ vs Fail)} \\
\midrule
Valence  & 1.0  & $1.73\times10^{-6}$ & 0.996 & 1.0 vs 3.8 \\
Arousal  & 81.0 & 0.0158              & 0.538 & 2.0 vs 3.0 \\
Control  & 88.0 & 0.00165             & $-0.645$ & 4.0 vs 2.5 \\
\bottomrule
\end{tabular}
\end{table}

\subsubsection{Condition-Level Differences}
\begin{figure}[t]
    \centering
    \includegraphics[width=\columnwidth]{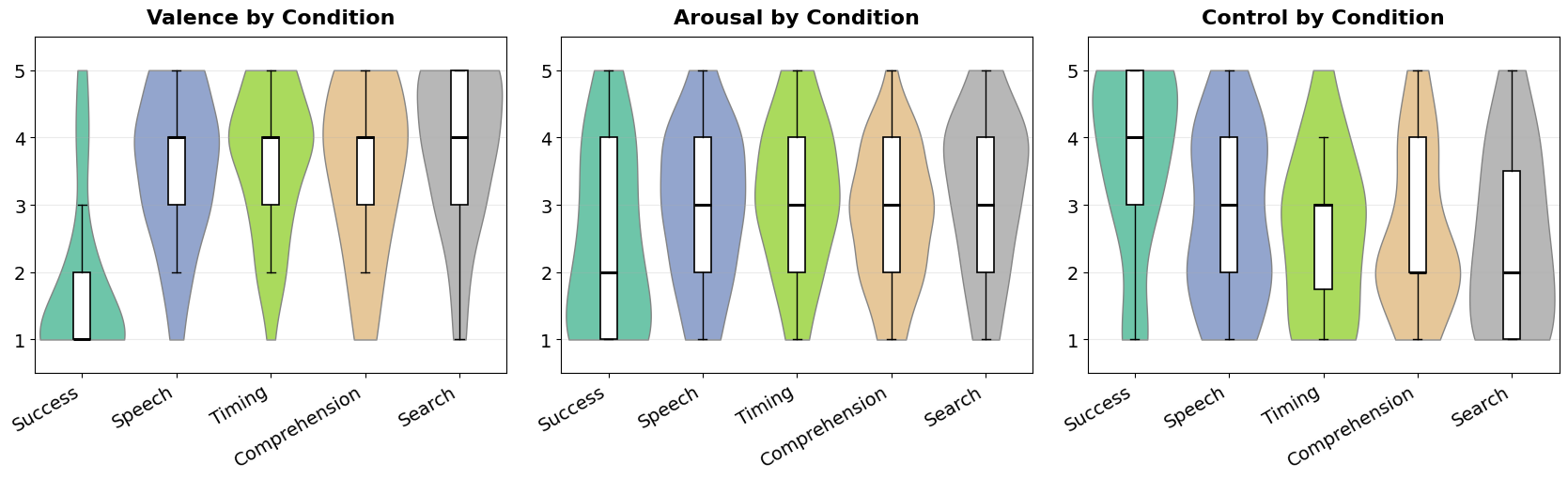}
    \caption{Affective ratings distribution across interaction conditions}
    \label{fig:affectxcnd}
\end{figure}

We conducted Friedman tests on $N = 28$ participants with complete data for all conditions ($k=5$). Significant differences were observed for valence, arousal, and control (Table~\ref{tab:affect_condition_tests}). Kendall’s $W$ indicated a strong effect for valence and small-to-moderate effects for arousal and control.
Descriptively, failure conditions were associated with higher valence ratings (more unpleasant experiences) and lower perceived control, whereas success trials were characterized by low valence ratings and high control (Figure \ref{fig:affectxcnd}). Median affective summaries by condition are reported in Table~\ref{tab:affect_condition_desc}.

\begin{table}[t]
\centering
\small
\caption{Friedman tests for affective differences across five interaction conditions ($n=28$ participants with complete data).}
\label{tab:affect_condition_tests}
\begin{tabular}{lcccc}
\toprule
\textbf{Dimension} & $\chi^2$ & \textbf{df} & \textbf{p-value} & \textbf{Kendall’s $W$} \\
\midrule
Valence  & 49.27 & 4 & $5.13\times10^{-10}$ & 0.440 \\
Arousal  & 11.20 & 4 & 0.0244               & 0.100 \\
Control  & 13.89 & 4 & 0.00765              & 0.124 \\
\bottomrule
\end{tabular}
\end{table}

\begin{table}[t]
\centering
\small
\caption{Median and interquartile range (IQR) of affective ratings by interaction condition.}
\label{tab:affect_condition_desc}
\setlength{\tabcolsep}{6pt}
\begin{tabular}{lccccccc}
\toprule
\textbf{Condition} & \textbf{$n$} &
\textbf{Valence} & &
\textbf{Arousal} & &
\textbf{Control} \\
 &  &
\textbf{Mdn} & \textbf{IQR} &
\textbf{Mdn} & \textbf{IQR} &
\textbf{Mdn} & \textbf{IQR} \\
\midrule
Comprehension Failure & 37 &
4.0 & [3.0--4.0] &
3.0 & [2.0--4.0] &
2.0 & [2.0--4.0] \\

Search Failure & 39 &
4.0 & [3.0--5.0] &
3.0 & [2.0--4.0] &
2.0 & [1.0--3.5] \\

Speech Failure & 41 &
4.0 & [3.0--4.0] &
3.0 & [2.0--4.0] &
3.0 & [2.0--4.0] \\

Timing Failure & 36 &
4.0 & [3.0--4.0] &
3.0 & [2.0--4.0] &
3.0 & [1.75--3.0] \\

Success & 36 &
1.0 & [1.0--2.0] &
2.0 & [1.0--4.0] &
4.0 & [3.0--5.0] \\
\bottomrule
\end{tabular}
\end{table}

\paragraph{Correlational Analyses}

Participant-level Spearman correlations (computed over median ratings across trials) revealed a moderate negative association between valence and control ($\rho=-0.436$) and between arousal and control ($\rho=-0.471$), while valence and arousal were largely uncorrelated ($\rho=-0.008$). These correlations suggest that more unpleasant or arousing interactions were associated with reduced perceived control.

\begin{figure}[t]
    \centering
    \includegraphics[width=\columnwidth]{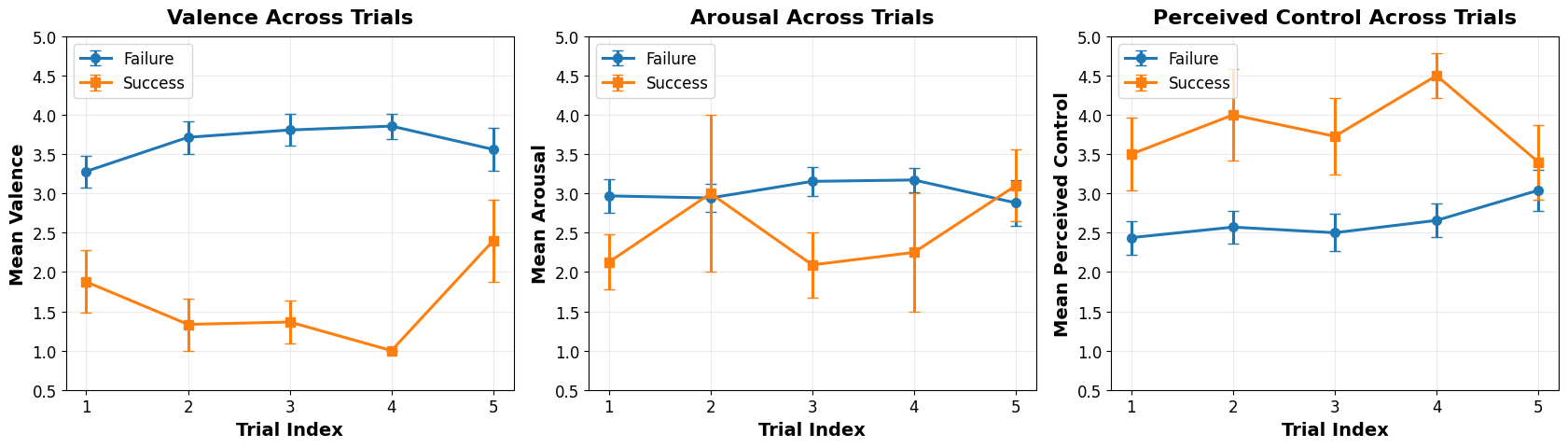}
    \caption{Affective ratings across trial order}
    \label{fig:affectxtrial}
\end{figure}

\paragraph{Temporal Analysis}
\label{Sec:affect_temporal}
Figure \ref{fig:affectxtrial} suggests no visible temporal variation. 

\subsection{Recovery Preferences}
\label{sec:results_recovery}

All included failure samples contained both the verbal and nonverbal recovery questions (i.e., no missing survey versions or files), yielding a fully eligible subset for modality-level summaries.

\subsubsection{Recovery Modality Distribution.} Across the eligible subset, recovery preferences were dominated by verbal strategies (Table~\ref{tab:recovery_modality}). Participants produced 431 verbal mentions (63.9\% of mentions) and 219 nonverbal mentions (32.5\%). Explicitly multimodal strategies (combined verbal + nonverbal in the same response) were rare (24 mentions; 3.6\%), and no responses indicated that recovery was unnecessary (0 mentions). 


\subsubsection{Strategy diversity.} Recovery preferences were heterogeneous, comprising 44 unique strategy strings overall, including 27 unique verbal strategies and 18 unique nonverbal strategies. The most frequently mentioned verbal strategies emphasized communicative clarity and transparency. Table~\ref{tab:recovery_top10_verbal} lists the ten most common verbal strategies, led by clarifying prompts (74 mentions) and self-correction/acknowledgment (59 mentions), followed by specific re-direction (45), transparency cues (42), and confidence checks (42).

\begin{table}[t]
\centering
\small
\caption{Recovery modality distribution (eligible subset; $N=153$ failure trials). Percentages are computed over all recovery mentions.}
\label{tab:recovery_modality}
\setlength{\tabcolsep}{8pt}
\begin{tabular}{lrr}
\toprule
\textbf{Modality} & \textbf{Mentions} & \textbf{Percent (\%)} \\
\midrule
Verbal      & 431 & 63.9 \\
Nonverbal   & 219 & 32.5 \\
Multimodal  & 24  & 3.6 \\
None needed & 0   & 0.0 \\
\midrule
\textbf{Total} & \textbf{674} & \textbf{100.0} \\
\bottomrule
\end{tabular}
\end{table}


\begin{table}[t]
\centering
\small
\caption{Top verbal recovery strategies by number of mentions (eligible subset; $N=153$ failure trials).}
\label{tab:recovery_top10_verbal}
\setlength{\tabcolsep}{6pt}
\begin{tabular}{p{0.5\columnwidth}r}
\toprule
\textbf{Verbal strategy (normalized label)} & \textbf{Mentions} \\
\midrule
Clarifying prompt (e.g., request repetition/clarification) & 74 \\
Self-correction \& acknowledgment                          & 59 \\
Specific re-direction (explicit drawer/item instruction)   & 45 \\
Transparency cue (state/explanation, e.g., recalibrating)  & 42 \\
Confidence check (explicit confirmation)                   & 42 \\
Guided reset (restart/close drawers and retry)             & 41 \\
Partial-understanding repair                               & 36 \\
Multimodal redundancy (speech paired with lights)          & 24 \\
Other (rare; long-tail responses)                          & 1 \\
\bottomrule
\end{tabular}
\end{table}

\section{Discussion}

\subsection{Comparison to Prior Work}

Our findings extend prior HRI failure research. First, while datasets like ERR@HRI 2.0 \cite{cao2025err} and REFLEX \cite{khanna2025reflex} examine conversational errors, we demonstrated that embodied item retrieval tasks produce distinct patterns, including confusion (30.7\%), frustration (19.1\%), and reduced perceived control (M=2.61 vs. M=3.61, p<.001). The negative correlation between valence and control ($\rho$ = -0.54, p < .001) indicates a deleterious effect of the loss of agency in robot failures.
Also, existing social signal detection work \cite{bremers2023socialcues, stiber2023using} recognizes failures but does not provide preferred recovery preferences for failure types. Our finding that 63.9\% of users prefer verbal transparency challenges the assumption that embodied robots primarily need physical recovery cues. Recovery preferences did not vary significantly by failure type ($\chi^2(27)=21.55$, p=.760), suggesting that universal principles (transparency, acknowledgment) outweigh strategies specific to failure types. 
While deployments \cite{taylor2025crashcart} identify real world robot failures, we demonstrate these can be systematically reproduced in controlled settings. Participants reported confusion as the primary emotion in several first trials but noted that this decreased over time, while frustration increased across trials. This trend confirms that repeated failures shift users from uncertain to irritated, providing quantitative evidence for RF-HIP predictions \cite{honig2018understanding}.

\subsection{Design Implications for Robot Failure Recovery}

Our findings suggest several implications for robot designers. First, robots should prioritize full transparency about failures. Participants consistently preferred recovery strategies that combined explicit verbal correction with transparency cues. Specifically, verbal strategies such as specific re-direction ("I meant syringes are in drawer 4") and self-correction acknowledgments ("Sorry, I made a mistake"), alongside non-verbal signals like lights blinking on the correct drawers or visual reset animations. This preference for multimodal, transparent recovery suggests that users want robots to clearly signal both \emph{that} an error occurred and \emph{how} to proceed.
Second, recovery strategies that include both verbal acknowledgment and actionable guidance are more effective than apologies alone. Non-verbal recovery cues (such as LED lights or visual animations) can complement verbal recovery but may be insufficient on their own, particularly for failures that leave users uncertain about what went wrong. This finding aligns with conversational repair literature \cite{alghamdi_system_2024} but extends it to physically grounded interaction.
Third, effective recovery may require matching the modality of the recovery strategy to the nature of the failure, though our data suggest this should be applied cautiously. 
Finally, user preferences may diverge from effective strategies in practice. That 63.9\% of preferences emphasized verbal recovery, despite the embodied item-retrieval task, suggests that users may underestimate multimodal redundancy. Our dataset enables testing whether preferences align with actual task performance.

\subsection{Limitations}

Our study has several limitations that constrain the generalizability of our findings. First, we used a Wizard-of-Oz protocol to ensure consistent failure injection across participants. While this enabled systematic data collection, it means our findings reflect user responses to simulated rather than real world robot failures. Autonomous failure detection and recovery systems may produce different user reactions, particularly if users perceive errors as reflecting genuine technical limitations (e.g., in terms of hardware and software) rather than experimental simulations.
Second, the crash cart robot was not deployed during real medical procedures, as the healthcare workers' interactions were collected alongside hospital corridors.
Third, our dataset captures only single-session interactions. We cannot assess how user reactions, trust, or recovery strategy preferences evolve with repeated exposure to robot failures over long term interactions. Longitudinal studies are needed to understand whether users adapt to characteristic failure types, whether social cues remain reliable indicators of failure perception over time, and how trust calibration unfolds in clinical roles where emotional restraint is normative.
Fourth, our sample of 41 participants, although including both healthcare workers and laypersons, limits our statistical power to detect interactions among failure type, user expertise, and recovery strategy preferences. 
Finally, our proposed failure modes focus on communication breakdowns (speech, timing, comprehension, search) and do not capture physical level errors such as cart navigation failures, collision avoidance errors, or integration failures with hospital information systems. These additional failure modes may elicit different user reactions and require different recovery strategies.

\subsection{Future Directions}
There are several potential directions from this work. First, \textbf{closing the loop from detection to recovery}: our dataset provides the multimodal signals needed to train classifiers that detect failure types from facial expressions, speech patterns, and behavioral cues, and to develop policies that select recovery strategies based on predicted failure type and user state. Because the dataset includes both objective behavioral traces and subjective recovery preferences, researchers can investigate whether user-preferred strategies align with objective efficacy, and can develop adaptive recovery systems that balance user preferences with task performance.

Second, \textbf{longitudinal trust calibration in clinical teams}: future work should examine how healthcare teams adapt to robots that make errors over extended deployments. Critical questions include whether repeated exposure to failures degrades or refines trust, how team coordination patterns shift when robots introduce uncertainty, and whether failure experiences lead to automation bias or appropriate reliance. Such studies would extend our single-session findings to understand the cumulative impact of robot failures on clinical workflow and team dynamics.

A third promising direction involves \textbf{autonomous failure management systems}: transitioning from Wizard-of-Oz simulation to fully autonomous robots that detect their own failures, infer user reactions, and execute context-appropriate recovery strategies. Our dataset provides a foundation for developing and benchmarking such systems, and future work involves validation of candidate recovery policies in real or high-fidelity clinical settings where failures have authentic consequences.

\section{Conclusion}

We presented RFM-HRI, a multimodal dataset that captures user reactions to systematically induce robot failures in a crash cart robot during an item retrieval task. Through Wizard-of-Oz studies with 41 participants across laboratory and hospital settings, we recorded synchronized facial expressions, head pose, speech, and self-reported responses to four failure types derived from three years of field observations. Our analysis revealed that confusion, annoyance, and frustration dominate user emotional responses to robot failures, with users consistently preferring recovery strategies that combine transparent verbal acknowledgment with actionable guidance. Recovery strategy preferences varied by failure type, suggesting that effective recovery may require matching recovery modalities to failure characteristics.

This work contributes to the HRI community by providing an empirically grounded set of failure modes for crash cart robots and robots created for item retrieval tasks, a publicly available multimodal benchmark dataset of robot failures with user reactions (RFM-HRI), and design insights for failure detection and recovery in physically grounded interactions. By capturing both the robot's technical perspective and the user's experiential perspective on failures, our dataset enables future research on adaptive recovery systems that can detect failures from multimodal behavioral signals and select context-appropriate recovery strategies. We hope this resource will advance the development of robots that not only acknowledge their limitations but actively work to repair interactions when failures inevitably occur.


\appendix

\pagebreak

\appendix
\counterwithin{figure}{section} 

\section{Wizard-of-Oz Interface Details}
\label{sec:appendix_woz}
In this appendix, we provide the complete visual breakdown of the Wizard-of-Oz interface used in both laboratory and hospital settings.

\begin{figure}[H]
    \centering
    \begin{subfigure}[b]{\columnwidth}
        \centering
        \includegraphics[width=\columnwidth]{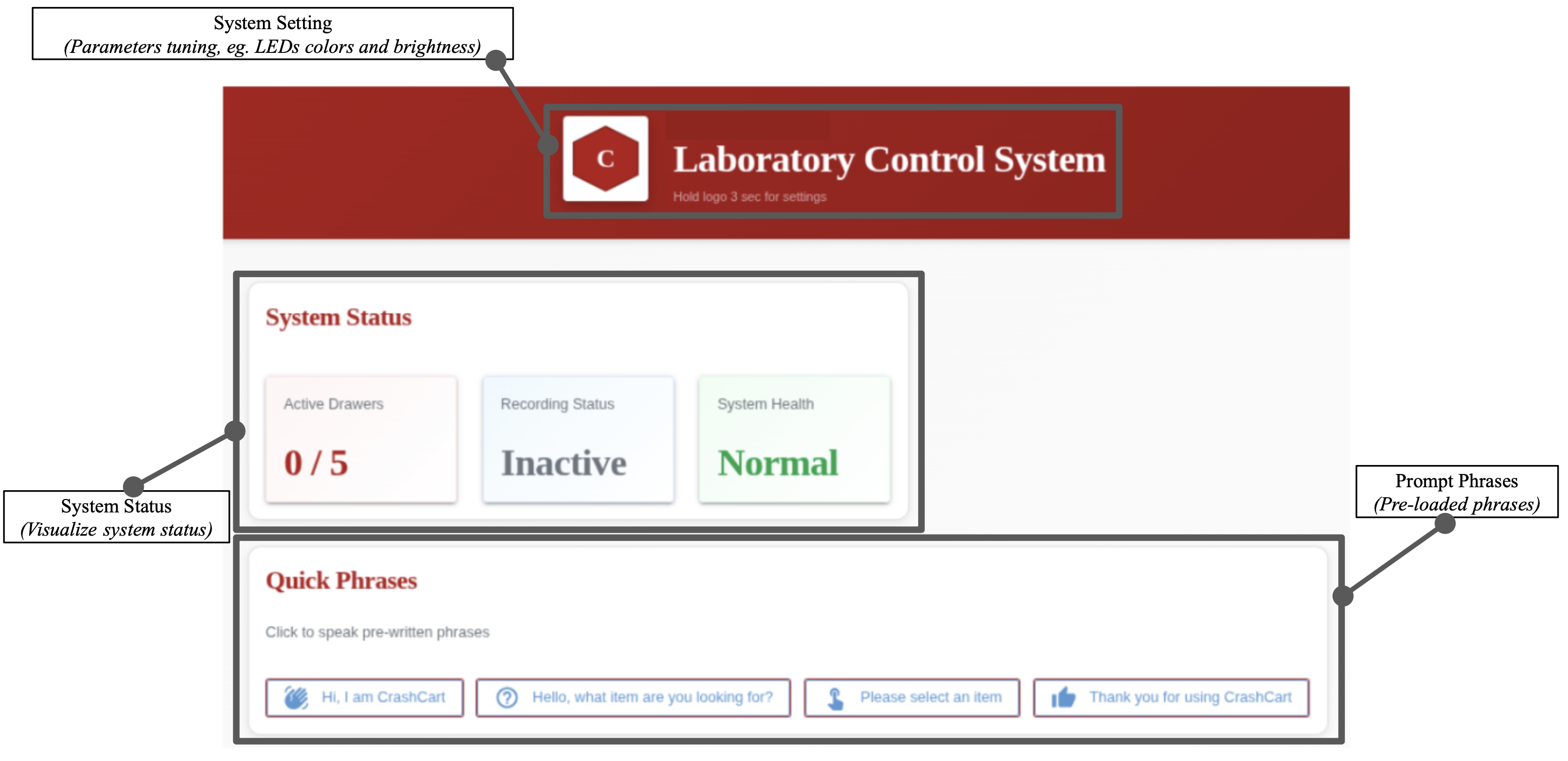}
        \caption{\textbf{Status \& Speech:} Monitors active drawers/health. "Quick Phrases" trigger pre-set TTS.}
        \label{fig:woz_status}
    \end{subfigure}

    \vspace{8pt}
    
    \begin{subfigure}[b]{\columnwidth}
        \centering
        \includegraphics[width=\columnwidth]{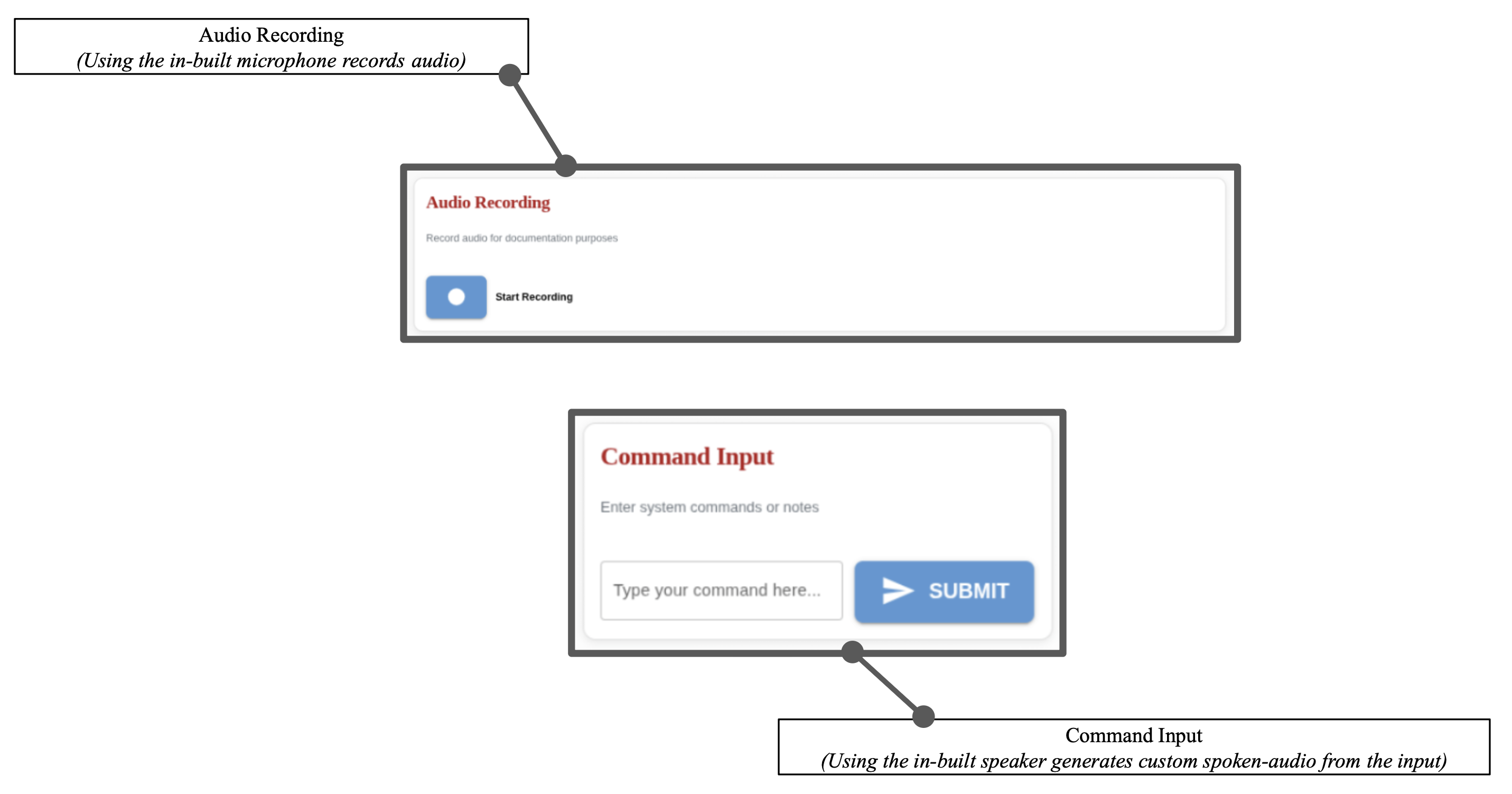}
        \caption{\textbf{Audio \& Input:} Recording controls and text input for custom TTS responses.}
        \label{fig:woz_audio}
    \end{subfigure}

    \caption{The Wizard-of-Oz (WoZ) Interface components (Part 3 of 4).}
    \label{fig:woz_part3}
\end{figure}

\begin{figure}[t!]
    \ContinuedFloat 
    \centering
    
    \begin{subfigure}[b]{\columnwidth}
        \centering
        \includegraphics[width=\columnwidth]{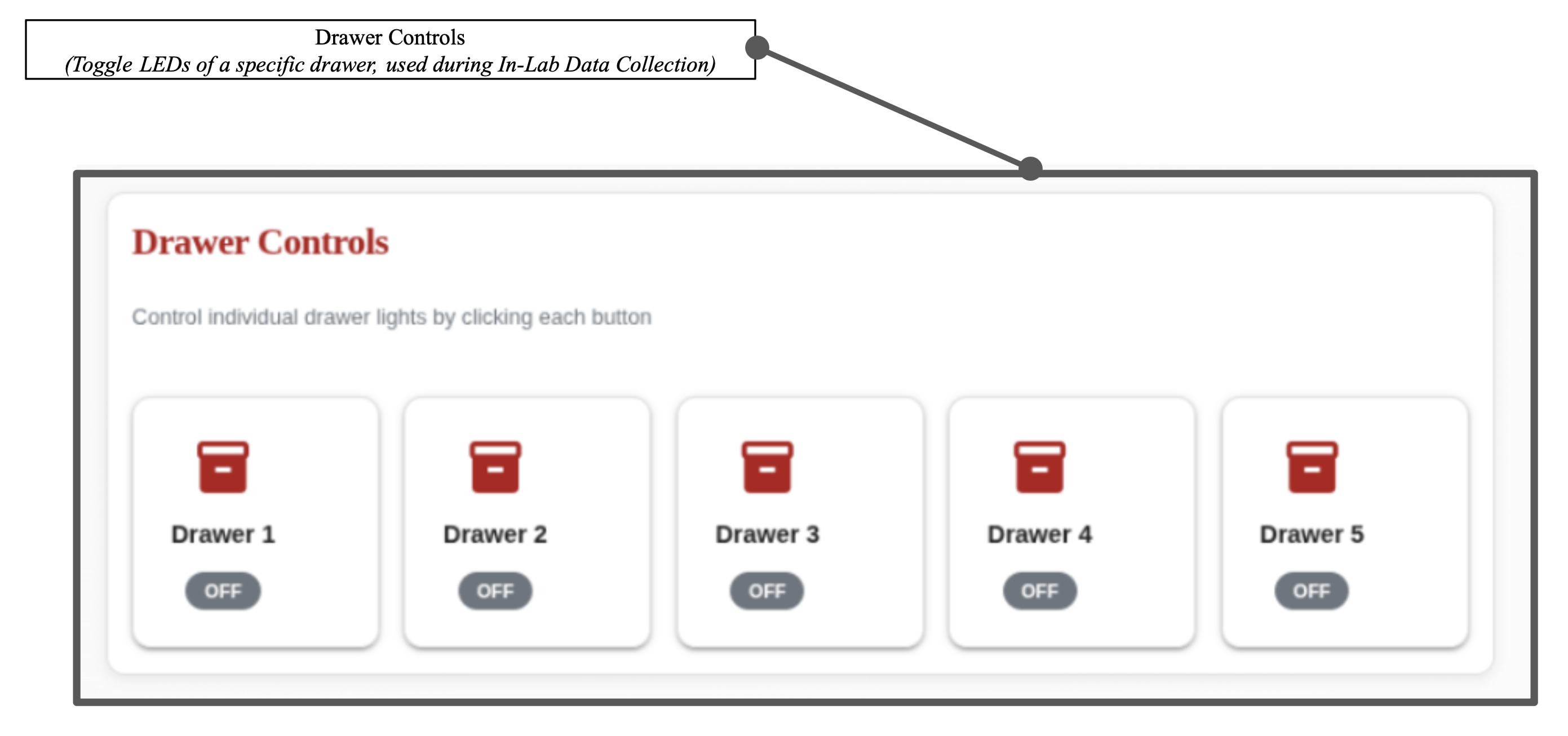}
        \caption{\textbf{Lab Mode (Manual):} The "Drawer Controls" module used for the in-lab proxy task.}
        \label{fig:woz_lab_manual}
    \end{subfigure}

    \vspace{8pt}
    
    \begin{subfigure}[b]{\columnwidth}
        \centering
        \includegraphics[width=0.9\columnwidth]{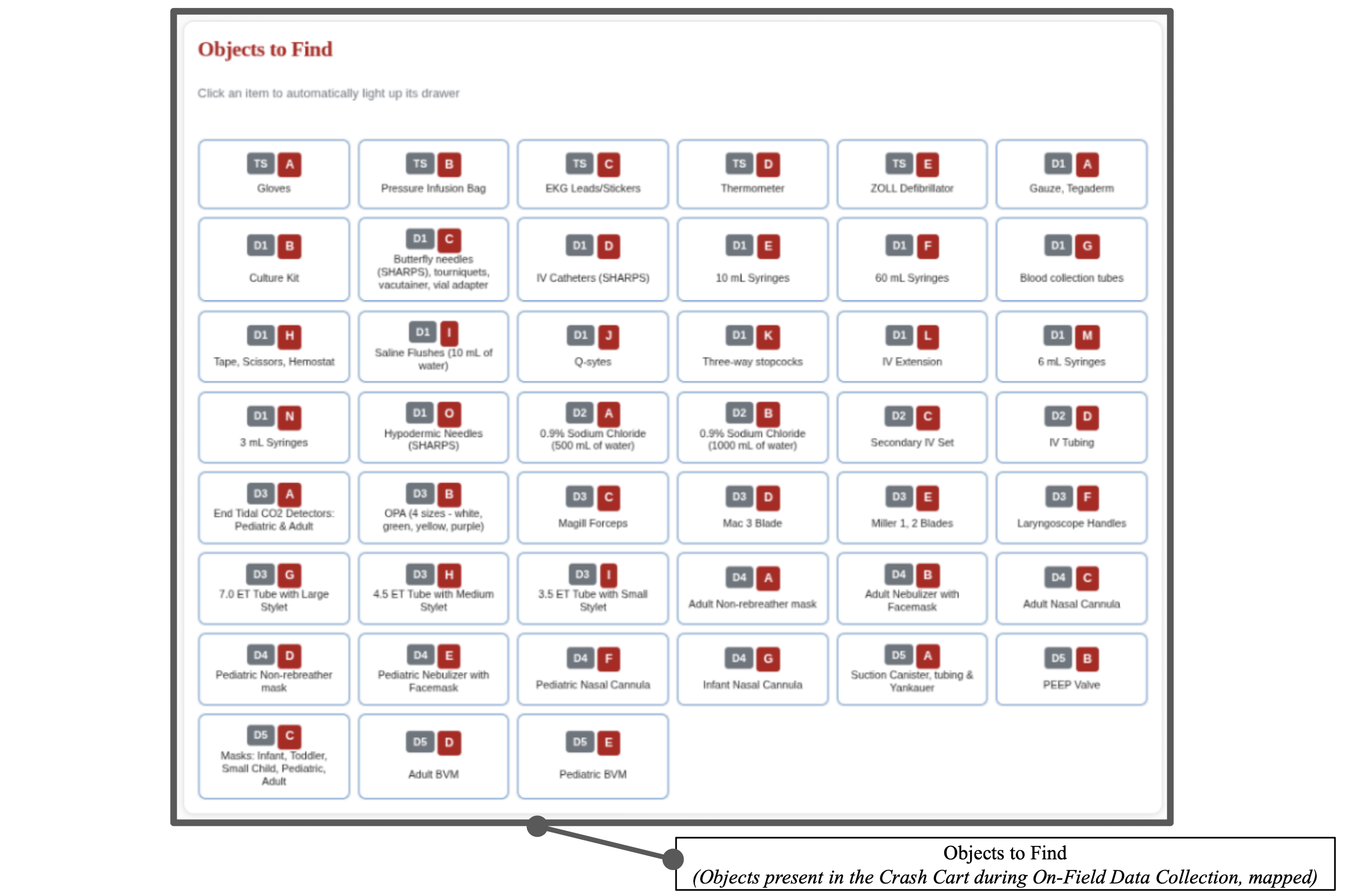} 
        \caption{\textbf{Hospital Mode (Mapped):} The "Objects to Find" grid used for the on-field medical task.}
        \label{fig:woz_hospital_grid}
    \end{subfigure}
    \caption{The Wizard-of-Oz (WoZ) Interface components (Part 4 of 4).}
    \label{fig:woz_part4}
\end{figure}

\end{document}